\documentclass[10pt,twocolumn,letterpaper]{article}

\usepackage[pagenumbers]{cvpr} 

\usepackage[dvipsnames]{xcolor}

\definecolor{cvprblue}{rgb}{0.21,0.49,0.74}
\usepackage[pagebackref,breaklinks,colorlinks,citecolor=cvprblue]{hyperref}

\newcommand*{\email}[1]{\texttt{#1}} %
\usepackage{multirow}
\usepackage{colortbl}

\definecolor{c_muns}{rgb}{0.88,1,1}
\definecolor{c_msup}{rgb}{1.0,0.88,1}
\definecolor{c_data}{rgb}{0.9,0.9,0.9}
\definecolor{c_lowbest}{rgb}{1.0,0.7,0.7}
\definecolor{c_highbest}{rgb}{0.7,0.7,1.0}
\definecolor{c_do}{rgb}{0.9,0.9,0.9}

\def\paperID{10945} 
\def\confName{CVPR}
\def\confYear{2024}

\title{From-Ground-To-Objects: Coarse-to-Fine Self-supervised Monocular Depth Estimation of Dynamic Objects with Ground Contact Prior}

\author{%
Jaeho Moon \quad\quad Juan Luis Gonzalez Bello \quad\quad Byeongjun Kwon \quad\quad Munchurl Kim\\
KAIST \\
\small{\email{\{jaeho.moon, juanluisgb, kbj2738, mkimee\}@kaist.ac.kr}} \\
\small{\url{https://kaist-viclab.github.io/From_Ground_To_Objects_site/}}
}

\begin{document}
\maketitle
\begin{abstract}

    Self-supervised monocular depth estimation (DE) is an approach to learning depth without costly depth ground truths.
    However, it often struggles with moving objects that violate the static scene assumption during training.
    To address this issue, we introduce a coarse-to-fine training strategy leveraging the ground contacting prior based on the observation that most moving objects in outdoor scenes contact the ground.
    In the coarse training stage, we exclude the objects in dynamic classes from the reprojection loss calculation to avoid inaccurate depth learning.
    To provide precise supervision on the depth of the objects, we present a novel Ground-contacting-prior Disparity Smoothness Loss (GDS-Loss) that encourages a DE network to align the depth of the objects with their ground-contacting points.
    Subsequently, in the fine training stage, we refine the DE network to learn the detailed depth of the objects from the reprojection loss, while ensuring accurate DE on the moving object regions by employing our regularization loss with a cost-volume-based weighting factor.
    Our overall coarse-to-fine training strategy can easily be integrated with existing DE methods without any modifications, significantly enhancing DE performance on challenging Cityscapes and KITTI datasets, especially in the moving object regions.
    
\end{abstract}

\vspace{-3mm}
\section{Introduction}
\label{sec:introduction}

Recent neural-network-based methods for depth estimation from 2D images have shown promising performance as the demand for 3D information grows with navigation, robotics, and AR/VR applications.
While supervised methods \cite{fu2018deep, lee2019big, ranftl2021vision, bhat2021adabins, li2023learning} utilize expensive and sparse ground truth (GT) depth for training, self-supervised methods learn depth by minimizing photometric errors (or reprojection loss) between the target frame and the warped frames from stereo images \cite{garg2016unsupervised, godard2017unsupervised, watson2019self, gonzalezbello2020forget, gonzalez2021plade, wang2023planedepth} or consecutive frames in monocular videos \cite{monodepth2, packing3d, temporal_opportunist, monovit}.

\begin{figure}
    \centerline{\includegraphics[width=0.45\textwidth]{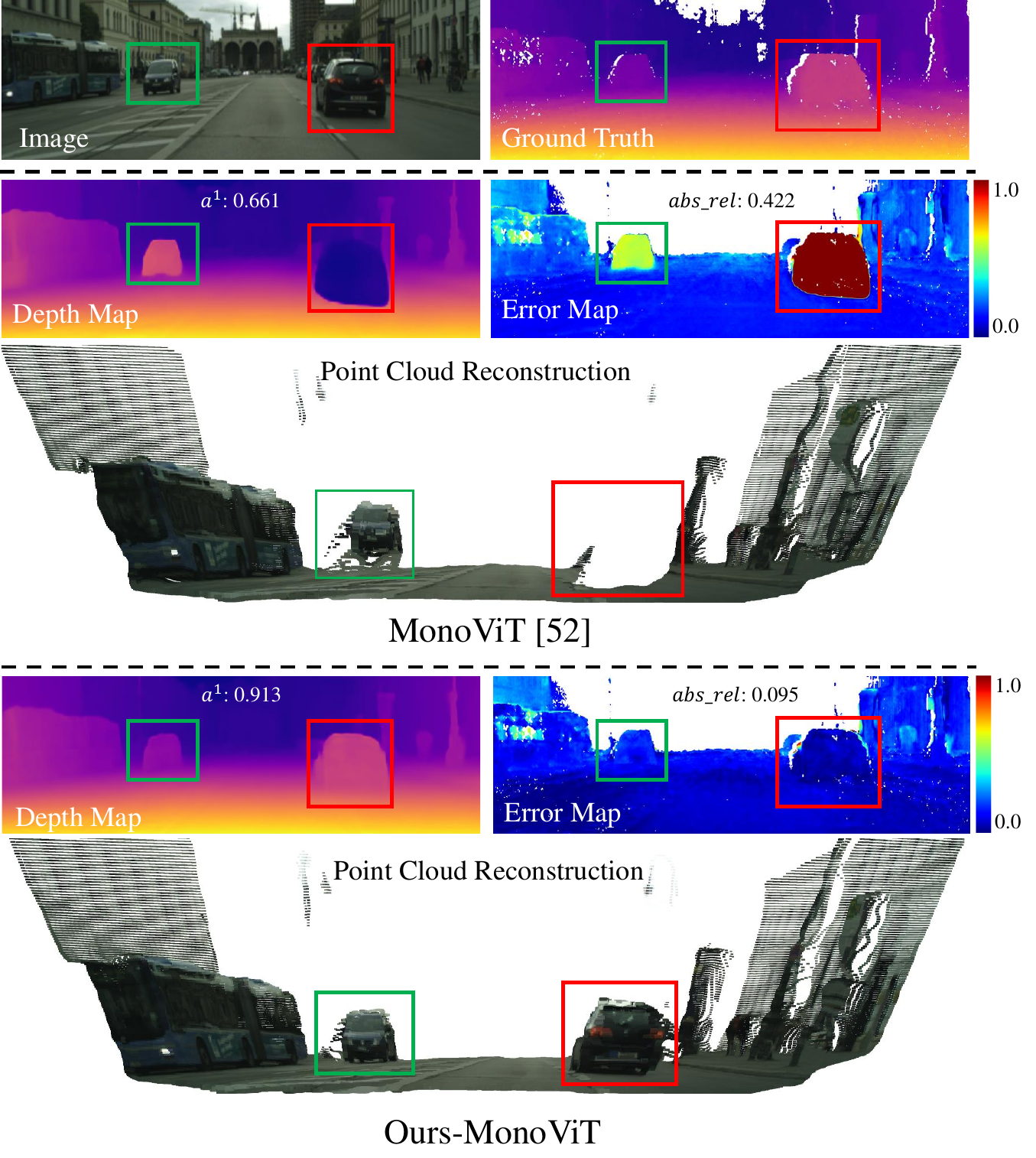}}
    \vspace*{-3mm}
    \caption{The effect of our coarse-to-fine self-supervised DE training strategy applied to MonoViT \cite{monovit}. In the error maps, blue indicates small errors, and red indicates large errors.} 
    \label{fig:figure_1}
\end{figure}

In the pipeline of self-supervised depth learning for monocular videos, a depth estimation (DE) network often learns inaccurate depth of moving objects that break the static scene assumption but still minimize the reprojection loss for self-supervision \cite{monodepth2, packing3d, semguide, klingner2020self}.
To handle the moving object problem, the `automasking' \cite{monodepth2} was proposed to exclude the static pixels in sequential frames from the reprojection loss calculation. 
However, automasking is limited to handling the objects moving at the same speed and direction as the camera. 
More recent work suggested to estimate 3D object motions \cite{li2021unsupervised, lee2021learning, lee2021attentive, rm_depth} or to disentangle object motion for cost volume construction \cite{feng2022disentangling}.
However, these self-supervised learning DE methods confront inherent ambiguities in learning the depth and motion of the objects without precise supervision.

In this paper, we introduce a ground contacting prior to effectively resolve the depth ambiguity of moving objects in a self-supervised manner.
The ground contacting prior is based on an observation that most objects classified as dynamic in outdoor scenes, such as cars, bicycles, or pedestrians, invariably tend to make contact with the ground, thereby sharing similar depth at their ground contact points.

Leveraging this crucial insight, we introduce a coarse-to-fine training strategy for self-supervised monocular DE.
In the initial coarse training stage, we train a DE network by employing the novel Ground-contacting-prior Disparity Smoothness Loss (GDS-Loss) that aligns the depth of objects in dynamic classes with the depth of their ground contact points.
Also, we exclude the objects from the reprojection loss computation to avoid learning inaccuracies in DE.
We further fine-tune the DE network in the fine training stage for learning to capture detailed depth over the surfaces of objects in dynamic classes by employing the reprojection loss for the whole image region.
We introduce a regularization loss that allows the DE network to learn the detailed depth of non-moving objects by the reprojection loss while avoiding inaccurate learning of DE on the moving object regions with a cost-volume-based weighting factor.

As shown in Fig. \ref{fig:figure_1}, MonoViT \cite{monovit} predicts inaccurate depth that make cars float (green box) or sunken under the ground (red box) in the reconstructed point cloud.
On the other hand, MonoViT trained with our coarse-to-fine training strategy, denoted as Ours-MonoViT, predicts the accurate depth of the cars, so that they are appropriately standing on the ground in the reconstructed point cloud. 
In summary, our contributions are three-fold:
\vspace{1mm}
\begin{itemize}
    \item We, for the first time, propose to utilize the ground contacting prior as a self-supervision for DE on objects in dynamic classes, presenting the novel Ground-contacting-prior Disparity Smoothness Loss (GDS-Loss). 
\vspace{1mm}
    \item We introduce a regularization loss with a cost-volume-based weighting factor, which allows fine-tunning with the reprojection loss while ensuring adherence to the ground contacting prior on the moving object regions.
\vspace{1mm}
    \item Our coarse-to-fine training strategy can easily be integrated into the existing DE networks, boosting their DE performance to achieve state-of-the-art results on the challenging Cityscapes and Kitti datasets.
\end{itemize}

\section{Related Work}
\label{sec:related_work}

\subsection{Self-supervised Monocular Depth Estimation}
Since measuring the GT depth is expensive, especially in outdoor environments, self-supervised learning of DE has been intensively studied \cite{garg2016unsupervised, godard2017unsupervised, sfmlearner, monodepth2, dijk2019neural, gonzalezbello2020forget, poggi2020uncertainty, temporal_opportunist, swami2022you, petrovai2022exploiting, han2022brnet}.
Without the GT depth, neural networks (NN) have been adopted to learn depth from stereo images \cite{poggi2018learning, pilzer2019refine, watson2019self, gonzalez2021plade, zhou2022self} or sequential video frames \cite{zhan2018unsupervised, chen2019self, shu2020feature, semguide, fsre_depth} by minimizing the photometric errors, or reprojection loss, between a target view and the warped reference view based on the predicted depth.
Since the usage of stereo images inherently has limitations such as requirements of synchronized binocular cameras and calibration, self-supervised learning from monocular videos for DE has been actively investigated. 

From monocular videos, NN-based methods learn from structure-from-motion techniques.
Zhou \textit{et al.} \cite{sfmlearner} incorporated two separate networks to learn camera ego-motion and depth simultaneously from monocular videos.
Godard \textit{et al.} proposed a CNN-based architecture, Monodepth2 \cite{monodepth2}, which handles occlusion problems based on the minimum reprojection loss.
Various approaches have been explored: ranging from improving network architectures, \cite{packing3d, hr_depth, johnston2020self, cadepth, monovit, bae2023deep}, using semantics-guidance \cite{klingner2020self, fsre_depth}, to employing sequential frames in test time by utilizing the cost volumes \cite{temporal_opportunist, guizilini2022multi, feng2022disentangling}.

Recently, Petrovai \textit{et al.} \cite{petrovai2022exploiting} suggested a two-stage training pipeline to transfer the knowledge of the high-resolution DE as explicit pseudo labels into the low-resolution DE training of a student network.
This process involves filtering out noisy pseudo depth labels by assessing 3D consistency across sequential frames.
In contrast, our coarse-to-fine training strategy guides a DE network to learn the depth of objects in dynamic classes based on the ground contacting prior (GDS-Loss) in the coarse stage and to further refine them in the fine stage.
Our approach employs a regularization loss in the fine stage to regularize the reprojection loss for enhancing the detailed depth predictions in static object regions while encouraging the precise learning of DE on the moving object regions with a cost-volume-based weighting factor.



\subsection{Handling Moving Objects}
Since the self-supervised monocular DE pipelines learn depth and camera ego-motion from consecutive frames under the rigid scene assumption, independently moving objects induce them to learn erroneous depth.
To overcome this problem, Monodepth2 \cite{monodepth2} utilized automasking that excludes pixels static in sequential frames, but it is only capable of masking out the objects moving at the same speed and in the same direction as the camera from the calculation of the reprojection loss.
Although other approaches suggested filtering out the images with moving objects from training scenes \cite{semguide}, and masking out the detected moving objects in photometric loss computation \cite{klingner2020self}, discerning whether an object is moving or not involves implicit uncertainty, which may lead to inaccurate DE performance.

Recent work \cite{gordon2019depth, li2021unsupervised, depth_wo_sensor, lee2021learning, lee2021attentive, feng2022disentangling, rm_depth} attempted to learn depth by decoupling object motion from camera ego-motion, exploiting a regularization term for 3D translation fields \cite{li2021unsupervised}, or predicting individual motions of objects by utilizing off-the-shelf segmentation models \cite{depth_wo_sensor, lee2021learning, lee2021attentive}.
Feng \textit{et al.} \cite{feng2022disentangling} suggested an occlusion-aware cost volume for multi-frame DE by adopting a cycle-consistency loss.
However, without precise supervision of depth or motion of moving objects, these methods often yield suboptimal results, since the reprojection loss can be minimized even with inaccurate depth and object motion predictions. 
In contrast, our proposed GDS-Loss provides precise depth supervision on the objects classified as dynamic classes, regardless of their moving speed and directions based on the ground point they stand, which is ground contacting prior.
Moreover, our method can easily be integrated with existing DE methods by utilizing instance segmentation masks only during training without necessitating of training an auxiliary object motion estimation network.

\begin{figure*}
    \centerline{\includegraphics[width=1\textwidth]{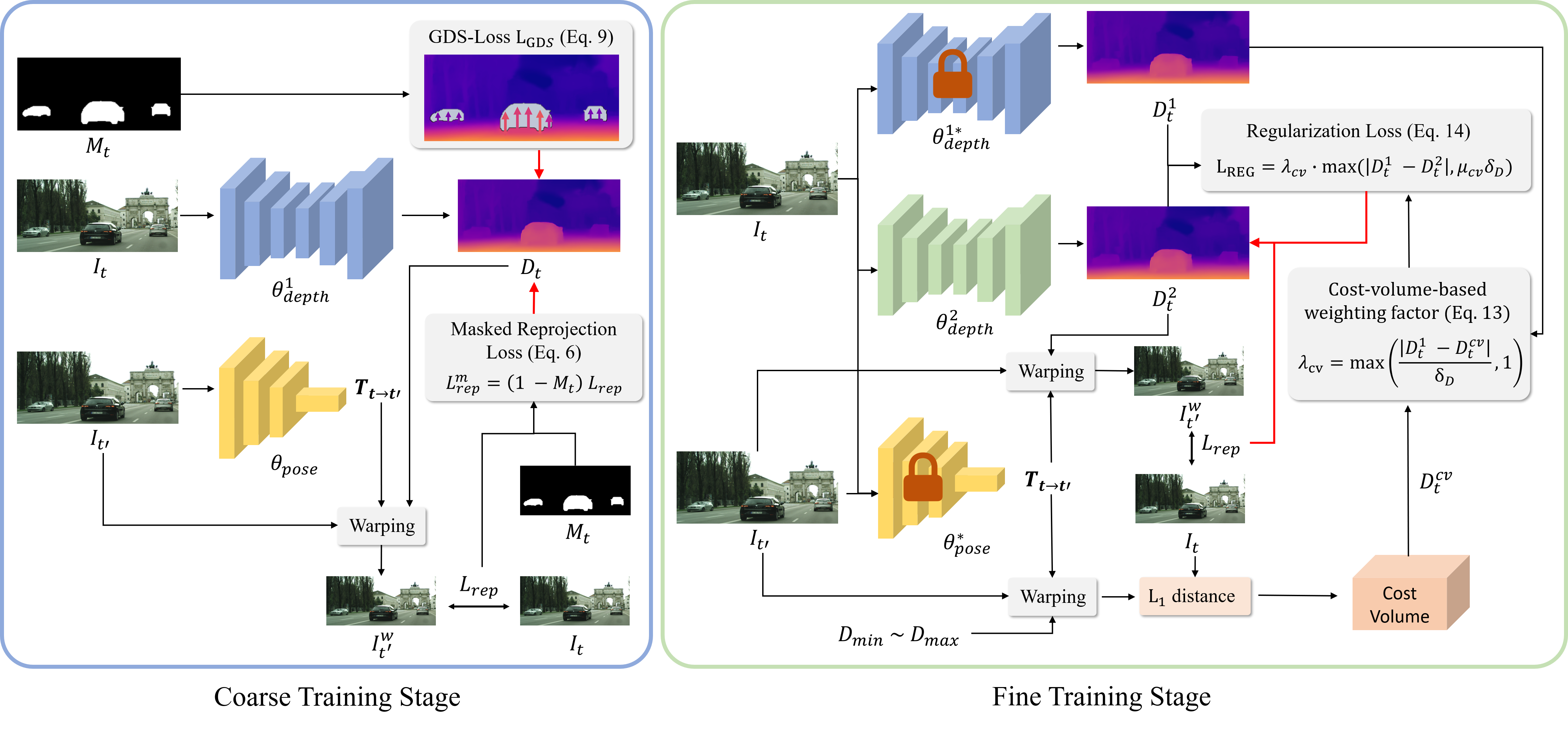}}
    \caption{Overview of our coarse-to-fine training strategy. In the coarse training stage, a DE network learns the depth of static region from the masked reprojection loss $L_{rep}^m$ and the depth of the objects in dynamic classes from our GDS-Loss $L_{GDS}$. In the fine training stage, we further refine the DE network with unmasked $L_{rep}$ while regularizing to ensure consistent depth prediction with the reliable DE network trained in the coarse stage by employing our regularization loss $L_{REG}$ with a cost volume-based weighting factor $\lambda_{cv}$.} 
    \label{fig:main_architecture}
\end{figure*}

\section{Method}
\label{sec:method}

\subsection{Self-supervised Monocular Depth Estimation}
\label{sec: ssmde_pipeline}

In this subsection, we briefly describe the conventional self-supervised monocular depth estimation (DE) pipeline in a self-contained manner.
Given two consequent frames $I_t$ and $I_{t'}$ at time $t$ and $t'$, respectively, from a monocular video, a DE network $\theta_{depth}$ predicts a depth map $D_t$ for target image $I_t$, as given by
\begin{equation}
    D_t = \theta_{depth} (I_t).
\end{equation}
Also, a pose estimation network $\theta_{pose}$ predicts a camera ego-motion $T_{t \rightarrow t'}$ from $I_t$ to $I_{t'}$.
We can obtain a synthesized frame $I_{t'}^w$ by warping $I_{t'}$ with the projected coordinates based on the estimated depth $D_t$, pose $T_{t \rightarrow t'}$ and the camera intrinsic matrix $K$, as given by
\begin{equation}
    I^w_{t'} = DW(I_{t'}, proj(D_t, T_{t\rightarrow t'}, K)),
\end{equation}
where $proj$ denotes the projection of the 3D transformed points and $DW$ indicates the differentiable warping process \cite{jaderberg2015spatial}. 
Then the reprojection loss $L_{rep}$ defined as the photometric errors between $I^w_{t'}$ and $I_t$, consisting of L1 and SSIM \cite{ssim} error terms weighted by $\alpha$, as given by
\begin{equation}
\label{eq:reprojection_loss}
    L_{rep} = \alpha \cdot L1(I_t, I^w_{t'}) + (1 - \alpha) \cdot SSIM(I_t, I^w_{t'}).
\end{equation}
In addition, an edge-aware smoothness loss $L_{sm}$ \cite{godard2017unsupervised} on the disparity $d_t$ (inverse of depth $D_t$) aims to encourage locally smooth depth maps while preserving the edges in the images, as is given by
\begin{equation}
\label{eq:disparity_smoothness_loss}
L_{sm} \:=\: |\partial_x \hat{d_t}|\:e^{-|\partial_x I_t|}\: + \:|\partial_y \hat{d_t}|\:e^{-|\partial_y I_t|},
\end{equation}
where $\hat{d_t}$ is the normalized disparity by the mean of $d_t$.
Specifically, each term in Eq. \ref{eq:disparity_smoothness_loss} consists of the disparity gradient exponentially weighted by the negative horizontal or vertical gradient of $I_t$.
As the color gradient $\partial_{x}I_t$ (or $\partial_{y}I_t$) decreases, $L_{sm}$ gets increased which guides $\theta_{depth}$ to predict smooth depth in homogeneous regions.
On the other hand, as $\partial_{x}I_t$ (or $\partial_{y}I_t$) increases, $L_{sm}$ decreases to induce abrupt boundaries in the depth map.
Through the training, $\theta_{depth}$ and $\theta_{pose}$ learn DE and camera ego-motion, respectively, by minimizing both $L_{rep}$ and $L_{sm}$, as given by
\begin{equation}
    L_{total} = L_{rep} + L_{sm}.
    \end{equation}

\subsection{Proposed Self-supervised DE Pipeline}
\label{sec: overall_training_strategy}

Fig. \ref{fig:main_architecture} illustrates our coarse-to-fine training strategy for self-supervised monocular DE. 
As depicted in the `Coarse Training Stage' (also in Sec. \ref{sec: coarse_stage}), DE network $\theta_{depth}^1$ learns the depth of objects in dynamic classes from our proposed Ground-contacting-prior Disparity Smoothness Loss $L_{GDS}$ (Eq. \ref{eq:GDS-Loss}) that provides precise supervision.
The region of objects in dynamic classes are masked out from the calculation of $L_{rep}$ (Eq. \ref{eq:masked_rep_loss}) to avoid inaccurate learning of DE.
After the `Coarse Training Stage', the trained DE network $\theta_{depth}^1$ and pose estimation network $\theta_{pose}$ are fixed and denoted as $\theta_{depth}^{1*}$ and $\theta_{pose}^{*}$, respectively. 
In the `Fine Training Stage' (also in Sec. \ref{sec: refinement_stage}), $\theta_{depth}^2$, initialized as the trained $\theta_{depth}^1$, is further finetuned to capture detailed depth over the surfaces of objects in dynamic classes by applying \textit{an unmasked} $L_{rep}$ on the whole image.
To prevent $\theta_{depth}^2$ from learning inaccurate DE on moving object regions, we introduce our regularization loss $L_{REG}$ (Eq. \ref{eq:reg_loss}) with a cost-volume-based weighting factor $\lambda_{cv}$ (Eq. \ref{eq:cost_volume_based_weight}).

\subsubsection{Coarse Training Stage with Ground Contacting Prior}
\label{sec: coarse_stage}

\textbf{Excluding objects in dynamic classes from $L_{rep}$ calculation.}
From the conventional self-supervised monocular DE pipeline in Sec. \ref{sec: ssmde_pipeline}, camera ego-motion is only considered for calculating $L_{rep}$.
Even though objects have independent motions that often differ from the camera, $L_{rep}$ can still be minimized via the estimated camera motion.
As a result, the DE network often fails to predict the precise depth of the objects, as shown in Fig. \ref{fig:figure_1}.
In this coarse training stage, instead of adopting erroneous supervision from $L_{rep}$, we mask objects that belong to dynamic classes, referred to as objects in dynamic classes, such as cars, bicycles, and pedestrians from calculating $L_{rep}$ whether the objects are moving or not.
Using a pre-computed instance segmentation map for input image $I_t$, we define the binary dynamic instance mask as $M_t$, valued 1 for the region of objects in dynamic classes and 0 otherwise.
The per-pixel masked reprojection loss $L_{rep}^m$ is defined as
\begin{equation}
\label{eq:masked_rep_loss}
    L_{rep}^{m} = (1 - M_t) L_{rep},
\end{equation}
so that the objects in dynamic classes can effectively be excluded from the calculation of $L_{rep}$, as also depicted in the `Coarse Training Stage' of Fig. \ref{fig:main_architecture}.

\vspace{1mm}
\noindent\textbf{Leveraging the ground contacting prior for supervision.}
A predominant cause of imprecise depth prediction for moving objects is insufficient supervision in the self-supervised learning of DE.
To address this issue, we introduce the ground contacting prior, rooted in the observation that most objects in dynamic classes such as cars, bicycles, and pedestrians, are in contact with the ground, resulting in similar depth with the depth of ground point they touch.
Based on this, we introduce a novel Ground-contacting-prior Disparity Smoothness Loss (GDS-Loss) that can precisely supervise the depth of objects in dynamic classes.
Our GDS-Loss is designed to guide DE networks for consistent depth predictions between the objects and their contacting ground locations.
As the ground is static, the DE networks can accurately predict the depth of the ground, which can be utilized for strong self-supervision on the objects.

\vspace{1mm}
\noindent\textbf{Ground-contacting-prior Disparity Smoothness Loss.}
Our GDS-Loss is implemented by extending the conventional edge-aware disparity smoothness loss (Eq. \ref{eq:disparity_smoothness_loss}).
Our GDS-Loss focuses on the vertical gradient of disparity $|\partial_y \hat{d_t}|$ to induce consistency of DE in the region between objects in dynamic classes and their contacting ground.
Formally, for $\hat{d_t} \in \mathbb{R}^{n\times m}$, $|\partial_y \hat{d_t}|$ represents the disparity difference between two vertically adjacent pixel locations, $(i,j)$ and its bottom neighbor $(i+1,j)$, which is expressed as:
\begin{equation}
    |\partial_y \hat{d_t}|(i, j) = |\hat{d_t}(i, j) - \hat{d_t}(i+1, j)|,  
\end{equation}
where $|\partial_y \hat{d_t}| \in \mathbb{R}^{(n-1) \times m}$.
In order to enforce consistency of depth between the objects in dynamic classes and their ground points of contact, we introduce a ground-contacting-prior mask $M_{gr}$ given as 
\begin{equation}
\label{eq:gravity_aware_mask}
    M_{gr}(i, j) = \gamma \cdot M_t(i, j) + (1 - M_t(i, j))
\end{equation}
where $M_{gr} \in \mathbb{R}^{(n-1) \times m}$ and $\gamma$ is a weighting parameter for $M_{gr}$, empirically setting to $100$.
For the location $(i, j)$ in the object region, we let $|\partial_y \hat{d_t}|$ be highly weighted by $M_{gr}(i, j) = \gamma$, so that the disparities are enforced to be consistent with their neighboring pixels in a downward direction.
On the other hand, $|\partial_y \hat{d_t}|$ is weighted the same as in Eq. \ref{eq:disparity_smoothness_loss} in the background region because of $M_{gr}(i, j) = 1$.
We additionally modify the color gradients $|\partial_y I_t|$ and $|\partial_x I_t|$ to guide consistent depth prediction inside the regions of objects in dynamic classes by replacing $I_t$ with $I^m_t = (1 - M_t) I_t$ to remove the effect from the color edges.
Our GDS-Loss is defined as:
\begin{equation}
\label{eq:GDS-Loss}
L_{GDS} \:=\: |\partial_x \hat{d_t}|\:e^{-|\partial_x I_t^m|}\:  + \:|\partial_y \hat{d_t}|\: M_{gr} e^{-|\partial_y I_t^m|},
\end{equation}
so that it induces smooth depth inside the object regions while aligning them with their contacting ground points.
Note that only $|\partial_y \hat{d_t}|$ is weighted by $M_{gr}$.

Finally, our total loss function $L_{total}^{1}$ in the coarse training stage with the masked reprojection loss and GDS-Loss is now defined by
\begin{equation}
\label{eq:total_loss_c}
    L_{total}^{1} = L_{rep}^{m} + \beta L_{GDS}.
\end{equation}

\begin{figure}
    \centerline{\includegraphics[width=0.48\textwidth]{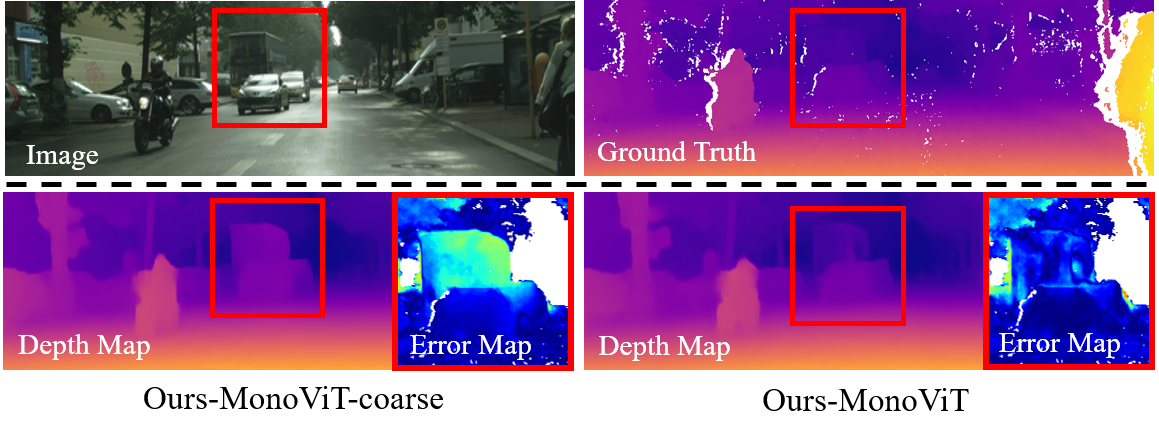}}
    \vspace*{-4mm}
    \caption{Our fine training stage further refines to capture the detailed depth of the objects in dynamic classes.} 
    \label{fig:coarse_to_fine}
\end{figure}

\subsubsection{Fine Training Stage}
\label{sec: refinement_stage}

\noindent\textbf{Refinement of the depth estimation network.}
The initial DE network $\theta_{depth}^1$, trained using the GDS-Loss (Eq. \ref{eq:GDS-Loss}) in the coarse training stage, effectively aligns the depth of the objects in dynamic classes with their ground contact points. 
Nonetheless, it may struggle to capture the detailed depth of objects when their surfaces vary vertically or if their bottom parts are occluded by other (closer to the camera) objects, leading to homogeneous depth prediction on both objects.
Thus, we introduce a fine training stage for $\theta_{depth}^2$ to learn the complex depth of the objects in dynamic classes by applying an $L_{rep}$ (Eq. \ref{eq:reprojection_loss}) for the whole image region. 
Note that $\theta_{depth}^2$ is initialized as the trained $\theta_{depth}^1$ weights and the fixed $\theta_{depth}^{1*}$ provides moderate supervision in this fine training stage.
Despite $L_{rep}$ is critical to induce DE networks to learn the detailed depth of non-moving objects, such as parked cars, that follow the rigid scene assumption, it is prone to guide $\theta_{depth}^2$ to learn the inaccurate DE on moving object regions, as discussed in Sec. \ref{sec: coarse_stage}.
 
\vspace{1mm}
\noindent\textbf{Regularization with respect to $\theta_{depth}^{1*}$.}
To regularize $\theta_{depth}^2$ from learning inaccurate DE, we propose to guide $\theta_{depth}^2$ to predict consistent depth with reliable depth predictions of $\theta_{depth}^{1*}$ that is trained based on the ground contacting prior.
We introduce a per-pixel depth regularization loss that guides $\theta_{depth}^2$ to learn detailed depth from $L_{rep}$ while maintaining alignment with $\theta_{depth}^{1*}$, which is expressed by
\begin{equation}
\label{eq:reg_loss_wo_factor}
    L_{REG}^0 = max(|D_t^1 - D_t^2|, \delta_{D}),
\end{equation}
where $D_t^1$ and $D_t^2$ represent the depth predictions of $\theta_{depth}^{1*}$ and $\theta_{depth}^2$ for target frame $I_t$, respectively.
$\delta_{D}$ is a hyperparameter for an allowable depth difference between $\theta_{depth}^2$ and $\theta_{depth}^{1*}$.
Note that $L_{REG}^0$ is activated only when the depth difference $|D_t^1 - D_t^2|$ exceeds $\delta_{D}$, while allowing $\theta_{depth}^2$ to learn fine details without regularization when $|D_t^1 - D_t^2|$ is minor.
However, our experiments indicate that the vanilla $L_{REG}^0$ fails to regularize when current $|D_t^1 - D_t^2|$ is small but quickly becomes larger by $L_{rep}$ in the moving object regions, so that $\theta_{depth}^2$ learns inaccurate DE.

\vspace{1mm}
\noindent\textbf{Cost-volume-based weighting factor.}
To enhance regularization in the moving object regions, we introduce a cost-volume-based approach that can identify the region by analyzing the depth discrepancy between $D_t^1$ and a cost-volume-induced depth. 
This approach is feasible because of the reliable depth prediction $D_t^1$ induced by the ground contacting prior.
The multi-view cost volume is constructed by the L1 distances between the target frame $I_t$ and the warped neighbor frame $I_t'$ to the position of $I_t$ with the depth candidates ranging from $D_{min}$ to $D_{max}$, following \cite{temporal_opportunist}. 
From the cost volume, the depth for each pixel that minimizes the L1 distance among depth candidates, which is the cost-volume-induced depth $D_{cv}$, is expressed as:
\begin{equation}
\label{eq:cost_volume_construction}
    D_t^{cv} = \underset{D_{k\in \{1, \dots, 32\}}}{\arg\min}\left(\left|DW(I_t', proj(D_k, T_{t \rightarrow t'}, K)) - I_t \right|\right),
\end{equation}
where $D_k$ represents a $k$-th quantized depth candidate in the range of [$D_{1} = D_{min}, D_{32} = D_{max}$].
$D_{min}$ and $D_{max}$ are determined by the minimum and maximum values of $D_t^1$, respectively. 
By leveraging the difference between the cost-volume-induced depth $D_t^{cv}$ and $D_t^1$, we define a pixel-wise cost-volume-based-weighting factor as
\begin{equation}
\label{eq:cost_volume_based_weight}
    \lambda_{cv} = max \left(\frac{|D_t^1 - D_t^{cv}|}{\delta_{D}}, 1 \right),
\end{equation}
where $\lambda_{cv}$ is equal to the normalized depth difference by $\delta_{D}$, being always greater than or equal to 1.
In order to regularize when $|D_t^1 - D_t^{cv}|$ is larger than $\delta_{D}$ in the moving object regions regardless of their current depth predictions, we additionally mask $\delta_{D}$ with $\mu_{cv} = [\lambda_{cv}=1]$ where $[\cdot]$ is an Iverson bracket.
By employing $\lambda_{cv}$ and $\mu_{cv}$ into $L_{REG}^0$ in Eq. \ref{eq:reg_loss_wo_factor}, our proposed pixel-wise regularization loss is defined as
\begin{equation}
\label{eq:reg_loss}
    L_{REG} = \lambda_{cv} \cdot max(|D_t^1 - D_t^2|, \mu_{cv} \delta_{D}),
\end{equation}
where $\delta_{D}$ is empirically set to 5\% of $D_{max}$ in our experiments.
The regions of moving objects have large $|D_t^1 - D_t^{cv}|$ values, so $L_{REG}$ regularizes the $\theta_{depth}^2$ learning toward maintaining the consistency between $D_t^1$ and $D_t^2$.
In contrast, the static regions which often have small values of $|D_t^1 - D_t^{cv}|$ and $|D_t^1 - D_t^{2}|$ can learn detailed depth without the regularization. 
Since $\lambda_{cv}$ penalizes more on the region that differs a lot between $D_{cv}$ and the reliable $D_t^1$, $L_{REG}$ is a more stable regularization than the approach that directly mask out pixels determined as moving from $L_{rep}$.
Our total loss $L_{total}^2$ in the fine training stage is defined as,
\begin{equation}
\label{eq:total_loss_f}
    L_{total}^{2} = L_{rep} + \rho L_{REG}.
\end{equation}

Note $L_{total}^{1}$ and $L_{total}^{2}$ are averaged over all image pixels and batches for training. As shown in Fig. \ref{fig:coarse_to_fine}, MonoViT \cite{monovit} trained with the coarse training stage predicts consistent depth between the bus and the front car in the red box, so the bus has a higher error than the car.
On the other hand, our fine training stage guides Our-MonoViT to predict the accurate depth of the occluded bus with low error.

\begin{table*}
\footnotesize
\centering
\begin{tabular}{ccccccccccccc}
\toprule
D & Methods & O.N. & Sem & Resolutions & $abs\:rel\downarrow\cellcolor{c_lowbest}$  & $sq\:rel\downarrow\cellcolor{c_lowbest}$ & $rmse\downarrow\cellcolor{c_lowbest}$  & $log_{rmse}\downarrow\cellcolor{c_lowbest}$ & $a^1\uparrow\cellcolor{c_highbest}$ & $a^2\uparrow$\cellcolor{c_highbest}  & $a^3\uparrow$\cellcolor{c_highbest}  \\ 
\midrule
\multirow{21}{*}{\rotatebox[origin=c]{90}{Cityscapes}} 
& Struct2Depth \cite{depth_wo_sensor} & \checkmark & Tr & $128 \times 416$ & 0.145 & 1.737 & 7.280 & 0.205 & 0.813 & 0.942 & 0.976 \\
& Gordon et al. \cite{gordon2019depth} & \checkmark & Tr & $128 \times 416$ & 0.127 & 1.330 & 6.960 & 0.195 & 0.830 & 0.947 & 0.981 \\
& Li et al. \cite{li2021unsupervised} & \checkmark & & $128 \times 416$ & 0.119 & 1.290 & 6.980 & 0.190 & 0.846 & 0.952 & 0.982 \\
& Monodepth2 \cite{monodepth2} & & & $128 \times 416$ & 0.129 & 1.569 & 6.876 & 0.187 & 0.849 & 0.957 & 0.983 \\
& \textbf{Ours-Monodepth2}&  & Tr & $128 \times 416$ &   0.110  &   1.179  &   6.390  &   0.169  &   0.881  &   0.968  &   0.989  \\
& DynamicDepth$\dagger$ \cite{feng2022disentangling} & & Tr/Te & $128 \times 416$ & 0.103 & 1.000 & 5.867 & 0.157 & 0.895 & 0.974 & 0.991 \\
& MonoViT \cite{monovit} & & & $128 \times 416$ & 0.114  &   1.238  &   6.589  &   0.174  &   0.860  &   0.965  &   0.990 \\
& \textbf{Ours-MonoViT} & & Tr &  $128 \times 416$ &   \textbf{0.096}  &   \textbf{0.930}  &   \textbf{5.806}  &   \textbf{0.152}  &   \textbf{0.905}  &   \textbf{0.976}  &   \textbf{0.992}  \\
\cmidrule{2-12}
& Lee et al. \cite{lee2021attentive} & \checkmark & Tr & $256 \times 832$ & 0.116 & 1.214 & 6.695 & 0.186 & 0.852 & 0.951 & 0.982 \\
& InstaDM \cite{lee2021learning} & \checkmark & Tr & $256 \times 832$ & 0.111 & 1.158 & 6.437 & 0.182 & 0.868 & 0.961 & 0.983 \\
& Monodepth2 \cite{monodepth2} & & & $192 \times 640$ & 0.125 & 1.474 & 6.688 & 0.180 & 0.865 & 0.964 & 0.988 \\
& \textbf{Ours-Monodepth2} & & Tr & $192 \times 640$ &   0.102  &   1.024  &   6.015  &   0.159  &   0.896  &   0.973  &   0.990  \\
& HR-Depth \cite{hr_depth} & & & $192 \times 640$ & 0.120  &   1.253  &   6.714  &   0.179  &   0.857  &   0.963  &   0.988  \\
& \textbf{Ours-HR-Depth} & & Tr & $192 \times 640$ & 0.100  & 1.010  &   5.998  &   0.157  &   0.896  &   0.974  &   0.991 \\
& RM-Depth \cite{rm_depth} & \checkmark & & $192 \times 640$ & 0.100 & 0.839 & 5.774 & 0.154 & 0.895 & 0.976 & 0.993 \\
& CADepth \cite{cadepth} & & & $192 \times 640$ & 0.124 & 1.278 & 6.771 & 0.183 & 0.862 & 0.962 & 0.986 \\
& \textbf{Ours-CADepth} & & Tr & $192 \times 640$ &   0.097  &   0.966  &   5.646  &   0.150  &   0.907  &   0.978  &   0.992  \\
& MonoViT \cite{monovit} & & & $192 \times 640$ & 0.106 & 1.098 & 6.071 & 0.160 & 0.881 & 0.974 & 0.991 \\
& \textbf{Ours-MonoViT} & & Tr & $192 \times 640$ &   \textbf{0.088}  &   \textbf{0.795}  &   \textbf{5.368}  &   \textbf{0.140}  &   \textbf{0.920}  &   \textbf{0.981}  &   \textbf{0.994}  \\
\midrule
\multirow{20}{*}{\rotatebox[origin=c]{90}{KITTI}}
& Struct2Depth \cite{depth_wo_sensor} & \checkmark & Tr & $192 \times 640$ & 0.141 & 1.026 & 5.291 & 0.215 & 0.816 & 0.945 & 0.979 \\
& Li et al. \cite{li2021unsupervised} & \checkmark & & $192 \times 640$ & 0.130 & 0.950 & 5.138 & 0.209 & 0.843 & 0.948 & 0.978 \\
& Gordon et al. \cite{gordon2019depth} & \checkmark & Tr & $192 \times 640$ &  0.128 & 0.959 & 5.230 & 0.212 & 0.845 & 0.947 & 0.976 \\
& SGDepth \cite{klingner2020self} & & Tr & $192 \times 640$ &  0.117 & 0.907 & 4.844 & 0.196 & 0.875 & 0.958 & 0.980 \\
& Lee et al. \cite{lee2021attentive} & \checkmark & Tr & $256 \times 832$ & 0.114 & 0.876 & 4.715 & 0.191 & 0.872 & 0.955 & 0.981 \\
& InstaDM \cite{lee2021learning} & \checkmark & Tr & $256 \times 832$  & 0.112 & 0.777 & 4.772 & 0.191 & 0.872 & 0.959 & 0.982 \\
& Monodepth2 \cite{monodepth2} & & & $192 \times 640$ & 0.115 & 0.903 & 4.863 & 0.193 & 0.877 & 0.959 & 0.981 \\
& \textbf{Ours-Monodepth2} & & Tr & $192 \times 640$ &    0.112  &   0.866  &   4.766  &   0.190  &   0.879  &   0.960  &   0.982  \\
& PackNet \cite{packing3d} & & & $192 \times 640$ & 0.111 & 0.785 & 4.601 & 0.189 & 0.878 & 0.960 & 0.982 \\
& Johnston et al. \cite{johnston2020self} & & & $192 \times 640$ & 0.111 & 0.941 & 4.817 & 0.189 & 0.885 & 0.961 & 0.981 \\
& HR-Depth \cite{hr_depth} & & & $192 \times 640$ & 0.109 & 0.792 & 4.632 & 0.185 & 0.884 & 0.962 & 0.983 \\
& \textbf{Ours-HR-Depth} & & Tr & $192 \times 640$ & 0.108 & 0.775 & 4.614 & 0.184 & 0.886 & 0.962 & 0.983 \\
& FSRE-Depth \cite{fsre_depth} & & Tr & $192 \times 640$ & 0.105 & 0.708 & 4.546 & 0.182 & 0.886 & 0.964 & 0.983 \\
& Petrovai et al. \cite{petrovai2022exploiting} & & & $192 \times 640$ & 0.106 & 0.751 & 4.485 & 0.180 & 0.885 & 0.964 & 0.984 \\
& RM-Depth \cite{rm_depth} & \checkmark & & $192 \times 640$ & 0.108 & 0.710 & 4.513 & 0.183 & 0.883 & 0.964 & 0.983 \\
& CADepth \cite{cadepth} & & & $192 \times 640$ & 0.105 & 0.769 & 4.535 & 0.181 & 0.892 & 0.964 & 0.983 \\
& \textbf{Ours-CADepth} & & Tr & $192 \times 640$ &  0.103  &   0.730  &   4.427  &   0.179  &   0.895  &   0.966  &   0.984  \\

& DynamicDepth$\dagger$ \cite{feng2022disentangling} & & Tr/Te & $192 \times 640$ & \textbf{0.096} & 0.720 & 4.458 & 0.175 & 0.897 & 0.964 & 0.984 \\
& MonoViT \cite{monovit} & & & $192 \times 640$ & 0.099 & 0.708 & 4.372 & 0.175 & 0.900 & 0.967 & 0.984 \\
& \textbf{Ours-MonoViT} & & Tr & $192 \times 640$ &   \textbf{0.096}  &   \textbf{0.696}  &   \textbf{4.327}  &   \textbf{0.174}  &   \textbf{0.904}  &   \textbf{0.968}  &   \textbf{0.985}  \\
\bottomrule
\end{tabular}   
\caption{Performance comparison of self-supervised DE methods on Cityscapes and KITTI. `D' column: Test dataset. $\dagger$: multiple test frames. A checkmark on the `O.N.': utilization of an additional object motion estimation network. `Tr'/`Te' in `Sem' column: semantic information usage in training/testing. Metrics with $\downarrow$ are the lower, the better, while metrics with $\uparrow$ are the higher, the better.}
\label{tab:main}
\end{table*}

\begin{figure*}
    \centerline{\includegraphics[width=0.95\textwidth]{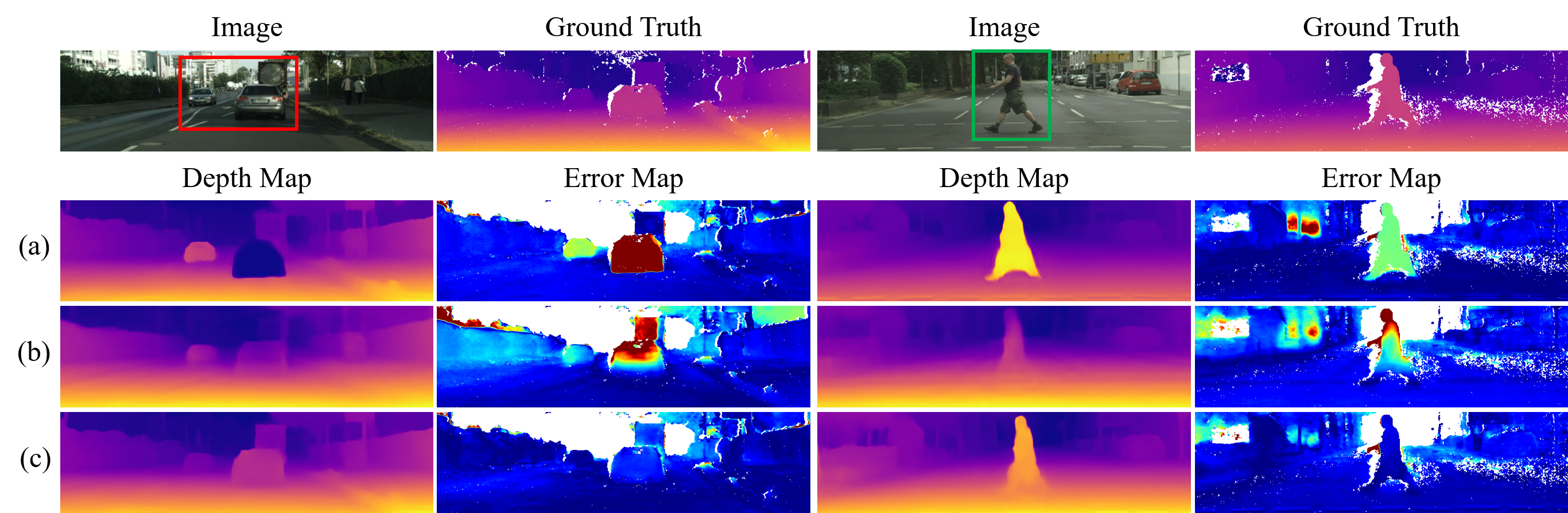}}
    \caption{Examples of estimated depth maps and error maps in Cityscapes \cite{cityscapes}. (a), (b), and (c) indicate MonoViT \cite{monovit}, DynamicDepth \cite{feng2022disentangling}, and Ours-MonoViT, respectively. In the error maps, blue indicates small errors, and red indicates large errors.} 
    \label{fig:cityscapes_result}
\end{figure*}

\section{Experiments}
\label{sec:experiments}

\subsection{Datasets}
Experiments are conducted on the widely adopted Cityscapes \cite{cityscapes} and KITTI \cite{kitti2015}.
Cityscapes \cite{cityscapes} contains a considerable amount of moving objects in the training dataset where self-supervised DE methods often suffer from the moving object problem.
Following \cite{sfmlearner,temporal_opportunist}, we use 69,731 training images (pre-processed with the scripts from \cite{sfmlearner}), and 1,525 testing images. 
For the DE performance on the regions of the objects in dynamic classes, we utilize masks provided by \cite{feng2022disentangling}.
We also train the DE networks on the KITTI \cite{kitti2015} dataset which contains 39,180 training and 4,424 validation images and evaluate on the Eigen split \cite{eigen_split} which contains 697 images.
For testing in both datasets, the estimated depth maps are normalized using median scaling within the depth range from 0\textit{m} to 80\textit{m}.

\subsection{Implementation Details}
\label{sec: implementation_details}
Our coarse-to-fine training strategy is integrated with the monocular DE networks including Monodepth2 \cite{monodepth2}, HR-Depth \cite{hr_depth} with ResNet-18 based encoder \cite{he2016deep}, CADepth \cite{cadepth} with ResNet-50 based encoder \cite{he2016deep} and MonoViT \cite{monovit} with MPViT-small based encoder, all pre-trained on ImageNet \cite{imagenet}.
The DE network and the pose estimation network (with ResNet-18 \cite{he2016deep} based encoder) are trained for 15 epochs in the coarse training stage and the DE network is refined for an additional 5 epochs in the fine training stage.
Our overall training strategy including the cost volume computation (Eq. \ref{eq:cost_volume_construction}) with the fixed $\theta_{depth}^{1*}$ in the fine stage increased the total training time by about 11.1\% (from 13.5h to 15h) on an NVIDIA RTX 3090 GPU for training Monodepth2 \cite{monodepth2} in Cityscapes.
The learning rates for ResNet-based models are set as $1 \times 10^{-4}$ for the coarse stage and $1 \times 10^{-5}$ for the fine stage, while Ours-MonoViT starts from $5 \times 10 ^{-5}$ and decaying exponentially by a factor of 0.9 for 20 epochs.
The overall pipeline is implemented in PyTorch \cite{paszke2019pytorch} and trained with the Adam \cite{adam} optimizer with a batch size of 12.
The loss weights $\alpha$ of the L1 loss (Eq. \ref{eq:reprojection_loss}), $\beta$ of the GDS-Loss (Eq. \ref{eq:total_loss_c}), and $\rho$ of $L_{REG}$ (Eq. \ref{eq:total_loss_f}) are empirically set to 0.85, 0.001 and 0.1, respectively.
The masks of objects in dynamic classes for training are pre-computed by the Mask2Former \cite{cheng2021mask2former} instance segmentation network with the backbone of Swin-S \cite{liu2021swin}, pre-trained from COCO \cite{coco} dataset. 

\begin{table}
\scriptsize
\setlength{\tabcolsep}{1.5pt} 
\begin{center}
    \begin{tabular}{cccccccc}
\toprule
Methods & Resolutions & $abs\:rel\downarrow\cellcolor{c_lowbest}$  & $sq\:rel\downarrow\cellcolor{c_lowbest}$ & $rmse\downarrow\cellcolor{c_lowbest}$  & $log_{rmse}\downarrow\cellcolor{c_lowbest}$ & $a^1\uparrow\cellcolor{c_highbest}$ \\ 
\midrule
Monodepth2 \cite{monodepth2} & $128 \times 416$ & 0.159 & 1.937 & 6.363 & 0.201 & 0.816 \\
MonoViT \cite{monovit} & $128 \times 416$ &  0.142  &   1.536  &   5.130  &   0.185  &   0.820  \\
Ours-Monodepth2 & $128 \times 416$ &   0.136  &   1.238  &   4.791  &   0.176  &   0.864   \\
DynamicDepth \cite{feng2022disentangling} & $128 \times 416$ & 0.129 & 1.273 & 4.626 & 0.168 & 0.862 \\
Ours-MonoViT & $128 \times 416$ &   \textbf{0.109}  &   \textbf{0.888}  &   \textbf{4.243}  &   \textbf{0.151}  &   \textbf{0.898}   \\
\midrule
InstaDM \cite{lee2021learning} & $256 \times 832$ & 0.139 & 1.698 & 5.760 & 0.181 & 0.859 \\
Monodepth2 \cite{monodepth2} & $192 \times 640$ & 0.185 & 2.432 & 5.919 & 0.218 & 0.794 \\
HR-Depth \cite{hr_depth} & $192 \times 640$ & 0.165 & 2.144 & 5.720 & 0.198 & 0.811 \\
CADepth \cite{cadepth} & $192 \times 640$ & 0.143 & 1.278 & 4.997 & 0.184 & 0.825 \\
MonoViT \cite{monovit} & $192 \times 640$ &  0.149  &   1.508  &   5.340  &   0.190  &   0.817  \\
Ours-Monodepth2 & $192 \times 640$ &   0.119  &   1.044  &   4.270  &   0.157  &   0.881   \\
Ours-HR-Depth & $192 \times 640$ &    0.117  &   1.015  &   4.297  &   0.154  &   0.883 \\
Ours-CADepth & $192 \times 640$ &   0.116  &   1.161  &   4.370  &   0.154  &   0.892  \\
Ours-MonoViT & $192 \times 640$ &   \textbf{0.098}  &   \textbf{0.674}  &   \textbf{3.688}  &   \textbf{0.135}  &   \textbf{0.914}   \\
\bottomrule
\end{tabular}
\end{center}
\vspace*{-3mm}
\caption{Performance comparison of self-supervised DE methods for objects in dynamic classes on Cityscapes \cite{cityscapes}.}
\label{tab:object}
\end{table}




\subsection{Cityscapes Results}

Table \ref{tab:main} shows the performance comparison of self-supervised DE methods on Cityscapes and KITTI datasets.
Especially on Cityscapes dataset which contains a substantial amount of moving objects, our training strategy significantly improves the DE performance of existing DE methods including Monodepth2 \cite{monodepth2}, HR-Depth \cite{hr_depth}, CADepth \cite{cadepth} and MonoViT \cite{monovit}, 
At a resolution of $192 \times 640$, the aforementioned methods are enhanced by incorporating our training strategy with large margins of 18.4\%, 16.7\%, 21.8\%, and 17.0\% in the $abs rel$ metric.
Particularly, Ours-Monodepth2 outperforms other methods \cite{hr_depth, cadepth, monovit} that are trained without our training strategy.
Also, Ours-CADepth and Ours-MonoViT outperform RM-Depth \cite{rm_depth} without requiring additional object motion estimation networks.
Notably, in the $128 \times 416$ resolution, our single-frame-based Ours-MonoViT outperforms the multi-frame-based SOTA DynamicDepth \cite{feng2022disentangling}, which relies on segmentation masks in test time.
In Table \ref{tab:object}, we compare the DE performance on the regions of objects in dynamic classes.
Our coarse-to-fine training strategy significantly enhances the DE performance of existing DE methods \cite{monodepth2, hr_depth, cadepth, monovit}.
Especially in a resolution of $128 \times 416$, Ours-MonoViT outperforms DynamicDepth \cite{feng2022disentangling} in all metrics.

Also, in Fig. \ref{fig:cityscapes_result}, MonoViT \cite{monovit} and DynamicDepth \cite{feng2022disentangling} predict erroneous depth on a car (denoted in a red box) and a pedestrian (denoted in a green box) resulting in high errors in the error maps due to the lack of supervision for depth of moving objects.
In contrast, Ours-MonoViT estimates accurate depth especially on the objects (the cars and the pedestrian) regardless of their moving speed and directions, while preserving details in depth maps.


\subsection{KITTI Results}

We further assess the efficacy of our coarse-to-fine training strategy on the frequently employed KITTI \cite{kitti2015} dataset.
As can be seen in Table \ref{tab:main}, the performances of Monodepth2 \cite{monodepth2}, HR-Depth \cite{hr_depth}, CADepth \cite{cadepth} and MonoViT \cite{monovit} are also improved with our training strategy integrated.
Although KITTI has less amount of moving objects in the training scenes, our coarse-to-fine training strategy consistently boost the DE performance of existing DE methods.
Also, Ours-MonoViT outperforms the current state-of-the-art multi-frame-based method, DynamicDepth \cite{feng2022disentangling}.

\section{Ablation Study}

We demonstrate the ablation study to verify the effectiveness of our coarse-to-fine training strategy by incrementally applying the proposed contributions to Monodepth2 \cite{monodepth2} and compare DE performance in Cityscapes \cite{cityscapes} dataset.

\begin{table}
\scriptsize
\setlength{\tabcolsep}{2pt} 
\begin{center}
    \begin{tabular}{ccc|c|cccccc}
\toprule
Exp  & $L_{rep}^{m}$ & $L_{GDS}$ &  & $abs\:rel\downarrow\cellcolor{c_lowbest}$  & $sq\:rel\downarrow\cellcolor{c_lowbest}$ & $rmse\downarrow\cellcolor{c_lowbest}$  & $log_{rmse}\downarrow\cellcolor{c_lowbest}$ & $a^1\uparrow\cellcolor{c_highbest}$ \\ 
\midrule
\multirow{2}{*}{(a)} & & & WIR & 0.125 & 1.474 & 6.688 & 0.180 & 0.865 \\
& & & DOR\cellcolor{c_do}  & 0.185\cellcolor{c_do} & 2.432\cellcolor{c_do} & 5.919\cellcolor{c_do} & 0.218\cellcolor{c_do} & 0.794\cellcolor{c_do}  \\
\midrule
\multirow{2}{*}{(b)} & \multirow{2}{*}{\checkmark} & & WIR & 0.119  &   1.492  &   6.610  &   0.179  &   0.877   \\
& & & DOR\cellcolor{c_do} & 0.160\cellcolor{c_do}  &   2.143\cellcolor{c_do}  &   5.503\cellcolor{c_do}  &   0.203\cellcolor{c_do}  &   0.813\cellcolor{c_do} \\
\midrule
\multirow{2}{*}{(c)}  & \multirow{2}{*}{\checkmark} & \multirow{2}{*}{\checkmark} & WIR & \textbf{0.104} & \textbf{1.097} & \textbf{6.034} & \textbf{0.160} & \textbf{0.895} \\
& & & DOR\cellcolor{c_do} & \textbf{0.125}\cellcolor{c_do} & \textbf{1.185}\cellcolor{c_do} & \textbf{4.459}\cellcolor{c_do} & \textbf{0.161}\cellcolor{c_do} & \textbf{0.876}\cellcolor{c_do}\\
\bottomrule
\end{tabular}
\end{center}
\vspace*{-3mm}
\caption{Ablation study on coarse training stage by training Monodepth2 \cite{monodepth2} on Cityscapes \cite{cityscapes}. WIR and DOR indicate the performance measures over the whole image regions and over the regions of objects in dynamic classes \textit{only}, respectively.}
\label{tab:ablation_coarse}
\end{table}

\vspace{2mm}
\noindent\textbf{Coarse Training Stage.}
In Table \ref{tab:ablation_coarse}, from Exp-(a) to (c), we train the DE network for 15 epochs to show the effectiveness of our proposed masked reprojection loss $L_{rep}^m$ and GDS-Loss $L_{GDS}$.

\vspace{1mm}
\textbf{Exp-(a)} - Original Monodepth2 \cite{monodepth2}: As shown in Table \ref{tab:ablation_coarse}, the automasking technique fails to guide the Monodepth2 \cite{monodepth2} model to learn the accurate depth of moving objects since it is only capable of excluding the objects moving at the same speed and direction as the camera.

\vspace{1mm}
\textbf{Exp-(b)} - Monodepth2 + $L_{rep}^m$: Simply excluding the objects in dynamic classes from the reprojection loss $L_{rep}$ shows marginal improvement by avoiding inaccurate learning of depth from $L_{rep}$. However, it is not sufficient to guide precise learning of depth due to the lack of supervision in the object regions.

\vspace{1mm}
\textbf{Exp-(c)} - Monodepth2 + $L_{rep}^m$ + $L_{GDS}$: Additional adoption of our GDS-Loss (Eq. \ref{eq:GDS-Loss}) not only enhances the overall performance but also drastically improves the DE quality on the regions of objects in dynamic classes, since our GDS-Loss $L_{GDS}$ provides precise depth supervision for the objects based on the ground contacting prior.

\begin{table}
\scriptsize
\setlength{\tabcolsep}{2pt} 
\begin{center}
    \begin{tabular}{cc|c|ccccc}
\toprule
Exp. & Methods & & $abs\:rel\downarrow\cellcolor{c_lowbest}$  & $sq\:rel\downarrow\cellcolor{c_lowbest}$ & $rmse\downarrow\cellcolor{c_lowbest}$  & $log_{rmse}\downarrow\cellcolor{c_lowbest}$ & $a^1\uparrow\cellcolor{c_highbest}$ \\ 
\midrule
\multirow{2}{*}{(d)}  & No & WIR &   0.145  &   2.868  &   7.395  &   0.201  &   0.856  \\
& regularization & DOR\cellcolor{c_do} &  0.303\cellcolor{c_do}  &  10.533\cellcolor{c_do}  &   8.839\cellcolor{c_do}  &   0.281\cellcolor{c_do}  &   0.764\cellcolor{c_do}  \\
\midrule
\multirow{2}{*}{(e)}  & \multirow{2}{*}{Filtering \cite{petrovai2022exploiting}}  & WIR &  0.124  &   1.620  &   6.739  &   0.187  &   0.868 \\
& & DOR\cellcolor{c_do} &  0.214\cellcolor{c_do}  &  4.372\cellcolor{c_do}  &  7.218\cellcolor{c_do}  &   0.257\cellcolor{c_do}  &   0.769\cellcolor{c_do} \\
\midrule
\multirow{2}{*}{(f)}  & \multirow{2}{*}{$L_{REG}^0$ (Eq. \ref{eq:reg_loss_wo_factor})} & WIR & 0.117  &   1.358  &   6.422  &   0.176  &   0.873   \\
&  & DOR\cellcolor{c_do} & 0.181\cellcolor{c_do}  &   2.594\cellcolor{c_do}  &   6.047\cellcolor{c_do}  &   0.218\cellcolor{c_do}  &   0.791\cellcolor{c_do} \\
\midrule
\multirow{2}{*}{(g)} &  \multirow{2}{*}{$L_{REG}$ (Eq. \ref{eq:reg_loss})} & WIR & \textbf{0.102} & \textbf{1.024} & \textbf{6.015} & \textbf{0.159} & \textbf{0.896}\\
&  & DOR\cellcolor{c_do} & \textbf{0.119}\cellcolor{c_do} & \textbf{1.044}\cellcolor{c_do} & \textbf{4.270}\cellcolor{c_do} & \textbf{0.157}\cellcolor{c_do} & \textbf{0.881}\cellcolor{c_do}  \\
\bottomrule
\end{tabular}
\end{center}
\vspace*{-3mm}
\caption{Ablation study on fine training stage by training Monodepth2 \cite{monodepth2} on Cityscapes \cite{cityscapes}. WIR and DOR indicate the performance measures over the whole image regions and over the regions of objects in dynamic classes \textit{only}, respectively.}
\label{tab:ablation_fine}
\end{table}

\vspace{2mm}
\noindent\textbf{Fine Training Stage.}
In Table \ref{tab:ablation_fine}, from Exp-(d) to (g), we show the effectiveness of our regularization loss $L_{REG}$ when the DE network $\theta_{depth}^2$, initially trained in Exp-(c), is further refined with the reprojection loss $L_{rep}$ to capture the detailed depth of objects in dynamic classes.

\vspace{1mm}
\textbf{Exp-(d)} - Training with no regularization: Although $\theta_{depth}^2$ is initially trained in the coarse stage, the DE performance shows degradation due to the inaccurate guidance from $L_{rep}$ in the moving object regions.

\vspace{1mm}
\textbf{Exp-(e)} - Usage of the filtering scheme in \cite{petrovai2022exploiting}: We mask out the pixels from $L_{rep}$, which are identified to be inconsistent in the 3D space by thresholding the difference between the depth of sequential frames predicted by $\theta_{depth}^{1*}$. The depth of neighbor frames are warped into the target frame position for consistency check. Although this approach is effective in filtering out noisy pixels from pseudo labels in \cite{petrovai2022exploiting}, it struggles to effectively exclude moving objects from $L_{rep}$ with thresholding so that $\theta_{depth}^2$ suffers from inaccurate learning from $L_{rep}$ unfiltered moving object regions.

\vspace{1mm}
\textbf{Exp-(f)} - Initial regularization loss $L_{REG}^0$ (Eq. \ref{eq:reg_loss_wo_factor}): Although $L_{REG}^0$ moderately guides $\theta_{depth}^2$ to predict consistent depth with $\theta_{depth}^1$, it struggles to effectively regularize the inaccurate supervision from $L_{rep}$, especially on the moving object regions.

\vspace{1mm}
\textbf{Exp-(g)} - Effectiveness of our regularization loss $L_{REG}$ (Eq. \ref{eq:reg_loss}) with $\lambda_{cv}$ (Eq. \ref{eq:cost_volume_based_weight}): Since $\lambda_{cv}$ effectively penalizes the moving object regions, $\theta_{depth}^2$ successfully improves the DE performance with elaborating the depth details of the objects in dynamic classes. Moreover, the approach of weighting based on the depth discrepancy relative to the reliable $\theta_{depth}^{1*}$ is shown to be smore stable than filtering pixels from $L_{rep}$ that involves implicit uncertainty by discerning whether an object is moving or not.

\section{Conclusion}
\label{sec:conclusion}

We address a challenging moving object problem in self-supervised monocular depth estimation.
For this, our proposed coarse-to-fine training strategy provides precise supervision on the depth of moving objects by firstly incorporating the \textit{ground-contacting-prior-based self-supervision}, which is Ground-contacting-prior Disparity Smoothness Loss (GDS-Loss).
The subsequent stage refines the DE network to capture the detailed depth of the objects in dynamic classes under well-defined constraints from our regularization loss with a cost-volume-based weighting factor.
The extensive experimental results show that our training strategy is very well harmonized with existing DE methods to effectively handle such moving objects.
Moreover, Ours-MonoViT yields the SOTA depth estimation performance on both KITTI and Cityscapes datasets.

\section*{Acknowledgements}
This work was supported by IITP grant funded by the Korea government (MSIT) (No. RS2022-00144444, Deep Learning Based Visual Representational Learning and Rendering of Static and Dynamic Scenes).


{
    \small
    \bibliographystyle{ieeenat_fullname}
    \bibliography{reference}
}

\clearpage

\setcounter{page}{1}
\maketitlesupplementary
\appendix

\section{Introduction}

In this supplementary material, we present an in-depth analysis of our proposed coarse-to-fine training strategy for self-supervised monocular depth estimation (DE). 
Specifically, we provide an extended ablation study focusing on two key components: the Ground-contacting-prior Disparity Smoothness Loss (GDS-Loss) employed in our coarse training stage (Sec. \ref{sec:supp_gds_loss}) and the regularization loss applied during our fine training stage (Sec. \ref{sec:supp_reg_loss}).

Additionally, we provide comprehensive experimental results showcasing the performance enhancements achieved by our training strategy on two benchmark datasets: Cityscapes \cite{cityscapes} (Sec. \ref{sec:supp_cityscapes_results}) and KITTI \cite{kitti2015} (Sec. \ref{sec:supp_kitti_results}). 
To further illustrate the impact of our approach, we present qualitative comparisons, including estimated depth maps and 3D point cloud reconstructions, to demonstrate the significant improvements brought by our coarse-to-fine training strategy (Sec.\ref{sec:supp_qualitative_results}).

\vspace{2mm}
\section{Additional Ablation Study}

\vspace{1mm}
\subsection{GDS-Loss}
\label{sec:supp_gds_loss}

Our proposed GDS-Loss $L_{GDS}$ induces a depth estimation (DE) network to align the depth of objects in dynamic classes (cars, bicycles, and pedestrians) to be consistent with the depth of their contacting ground.
For this, we extend the edge-aware disparity smoothness loss \cite{godard2017unsupervised} with the ground-contacting-prior mask $M_{gr}$ given as
\begin{equation}
\label{eq:supp_gravity_aware_mask}
    M_{gr}(i, j) = \gamma \cdot M_t(i, j) + (1 - M_t(i, j)),
\end{equation}
where $M_t \in \mathbb{R}^{(n-1) \times m}$ is a binary dynamic instance mask, valued 1 for the region of objects in dynamic classes and 0 otherwise.
We utilize $M_{gr}$ in our GDS-Loss $L_{GDS}$, which is defined as
\begin{equation}
\label{eq:supp_GDS-Loss}
L_{GDS} \:=\: |\partial_x \hat{d_t}|\:e^{-|\partial_x I_t^m|}\:  + \:|\partial_y \hat{d_t}|\: M_{gr} e^{-|\partial_y I_t^m|},
\end{equation}
where input image $I_t$ is replaced with the masked image $I_t^m$ for guiding consistent depth inside the object regions.
A higher value of $\gamma$ in Eq. \ref{eq:supp_gravity_aware_mask} encourages a DE network to predict consistent depth between the objects in dynamic classes and their contacting ground points.

\begin{table}
\scriptsize
\setlength{\tabcolsep}{1.5pt} 
\begin{center}
    \begin{tabular}{c|c|cccccc}
\toprule
$\gamma$ in $L_{GDS}$ &  & $abs\:rel\downarrow\cellcolor{c_lowbest}$  & $sq\:rel\downarrow\cellcolor{c_lowbest}$ & $rmse\downarrow\cellcolor{c_lowbest}$  & $log_{rmse}\downarrow\cellcolor{c_lowbest}$ & $a^1\uparrow\cellcolor{c_highbest}$ \\ 
\midrule
\multirow{2}{*}{1} & WIR & 0.113  &   1.404  &   6.349  &   0.169  &   0.890  \\
 & DOR\cellcolor{c_do}  &  0.159\cellcolor{c_do} & 2.353\cellcolor{c_do} & 5.825\cellcolor{c_do} & 0.198\cellcolor{c_do} & 0.844\cellcolor{c_do}  \\
\midrule
\multirow{2}{*}{10} & WIR & 0.110  &   1.262  &   6.218  &   0.166  &   0.888   \\
 & DOR\cellcolor{c_do} & 0.152\cellcolor{c_do}  &   2.023\cellcolor{c_do}  &   5.451\cellcolor{c_do}  &   0.190\cellcolor{c_do}  &   0.849\cellcolor{c_do} \\
\midrule
\multirow{2}{*}{100} & WIR & \textbf{0.104} & \textbf{1.097} & \textbf{6.034} & \textbf{0.160} & \textbf{0.895} \\
 & DOR\cellcolor{c_do} & \textbf{0.125}\cellcolor{c_do} & \textbf{1.185}\cellcolor{c_do} & \textbf{4.459}\cellcolor{c_do} & \textbf{0.161}\cellcolor{c_do} & \textbf{0.876}\cellcolor{c_do}\\
\midrule
\multirow{2}{*}{1000} & WIR & 0.105  &   1.140  &   6.134  &   0.164  &   0.889   \\
 & DOR\cellcolor{c_do} & 0.131\cellcolor{c_do}  &   1.044\cellcolor{c_do}  &   4.545\cellcolor{c_do}  &   0.166\cellcolor{c_do}  &   0.850\cellcolor{c_do} \\
\bottomrule
\end{tabular}
\end{center}
\caption{Ablation study on the value of $\gamma$ in our GDS-Loss $L_{GDS}$ by training Monodepth2 \cite{monodepth2} on Cityscapes \cite{cityscapes} in a resolution of $192 \times 640$. WIR and DOR indicate the performance measures over the whole image regions and over the regions of objects in dynamic classes \textit{only}, respectively.}
\label{tab:supp_gamma}
\end{table}

Table \ref{tab:supp_gamma} presents a comparative analysis of DE performance, influenced by varying $\gamma$ values ([1, 10, 100, 1000]) in GDS-Loss.
Please note that we train Monodepth2 \cite{monodepth2} with masking the objects in dynamic classes from the reprojection loss $L_{rep}$ for all experiments in Table \ref{tab:supp_gamma} to avoid inaccurate learning on the object regions. 
When $\gamma$ is low (1 or 10), the ground-contacting-prior mask $M_{gr}$ fails to effectively guide the DE consistency between the objects and their contacting ground, leading to suboptimal DE performance, especially on the dynamic class object regions. 
It is shown that a value of 100 for $\gamma$ achieves the best DE performance in both whole image regions and the regions of objects in dynamic classes.

\begin{table}
\scriptsize
\setlength{\tabcolsep}{1pt} 
\begin{center}
    \begin{tabular}{c|c|c|cccccc}
\toprule
Filtering & $L_{REG}$ &  & $abs\:rel\downarrow\cellcolor{c_lowbest}$  & $sq\:rel\downarrow\cellcolor{c_lowbest}$ & $rmse\downarrow\cellcolor{c_lowbest}$  & $log_{rmse}\downarrow\cellcolor{c_lowbest}$ & $a^1\uparrow\cellcolor{c_highbest}$ \\ 
\midrule
\multirow{2}{*}{\cite{petrovai2022exploiting}} & & WIR &  0.124  &   1.620  &   6.739  &   0.187  &   0.868 \\
& & DOR\cellcolor{c_do} &  0.214\cellcolor{c_do}  &  4.372\cellcolor{c_do}  &  7.218\cellcolor{c_do}  &   0.257\cellcolor{c_do}  &   0.769\cellcolor{c_do} \\
\midrule
& \multirow{2}{*}{$\frac{\delta_D}{D_{max}}=0.01$} &  WIR & 0.106 & 1.072 & 6.230 & 0.165 & 0.883 \\
& & DOR\cellcolor{c_do}  & 0.120\cellcolor{c_do} & 1.054\cellcolor{c_do} & 4.377\cellcolor{c_do} & 0.160\cellcolor{c_do} & 0.875\cellcolor{c_do}  \\
\midrule
& \multirow{2}{*}{$\frac{\delta_D}{D_{max}}=0.05$} & WIR & \textbf{0.102} & \textbf{1.024} & \textbf{6.015} & \textbf{0.159} & \textbf{0.896}\\
& & DOR\cellcolor{c_do} & \textbf{0.119}\cellcolor{c_do} & \textbf{1.044}\cellcolor{c_do} & \textbf{4.270}\cellcolor{c_do} & \textbf{0.157}\cellcolor{c_do} & \textbf{0.881}\cellcolor{c_do}  \\
\midrule
& \multirow{2}{*}{$\frac{\delta_D}{D_{max}}=0.1$} & WIR & 0.103 & 1.109 & 6.074 & 0.159 & 0.896 \\
& & DOR\cellcolor{c_do} & 0.120\cellcolor{c_do} & 1.112\cellcolor{c_do} & 4.379\cellcolor{c_do} & 0.157\cellcolor{c_do} & 0.881\cellcolor{c_do}\\
\bottomrule
\end{tabular}

\end{center}
\caption{Ablation study on the fine training stage by finetuning Monodepth2 \cite{monodepth2} initially trained with our coarse training stage on Cityscapes \cite{cityscapes} in a resolution of $192 \times 640$. WIR and DOR indicate the performance measures over the whole image regions and over the regions of objects in dynamic classes \textit{only}, respectively.}
\label{tab:supp_delta}
\end{table}

\vspace{2mm}
\subsection{Regularization Loss}
\label{sec:supp_reg_loss}

Following the initial training using GDS-Loss $L_{GDS}$ and a masked reprojection loss $L_{rep}^m$ in the coarse stage, we refine the depth estimation (DE) network. 
We denote the fixed DE network that is initially trained in the coarse stage as $\theta_{depth}^{1*}$ and the DE network to be further refined as $\theta_{depth}^{2}$.
Please note that the DE network $\theta_{depth}^2$ is initialized as the weight of $\theta_{depth}^{1*}$.
In Table \ref{tab:supp_delta}, we present the DE performance of the Monodepth2 \cite{monodepth2} model trained with various regularizations in the fine training stage.

In the first row of Table \ref{tab:supp_delta}, the DE performance of the regularization using the filtering scheme proposed in \cite{petrovai2022exploiting} is shown.
This method utilizes the depth predictions of $\theta_{depth}^{1*}$ for the target frame and its neighbor frames, while The depth maps of neighbor frames are warped into the position of the target frame with the scale adjustment.
The 3D inconsistent pixels with large depth differences are identified based on the threshold and excluded from the calculation of the reprojection loss $L_{rep}$.
However, this filtering scheme sometimes fails to filter out pixels and induces inaccurate learning of depth in the moving object regions.
As a result, it is shown that the DE performance of the DE network regularized with the filtering scheme \cite{petrovai2022exploiting} is degraded, especially in the region of objects in dynamic classes.

In contrast, our proposed regularization loss focuses on aligning the depth predictions of $\theta_{depth}^2$ with those of $\theta_{depth}^{1*}$, instead of excluding pixels from the reprojection loss $L_{rep}$.
In order to penalize the pixels in the moving object regions, we utilize the cost-volume-based-weighting factor $\lambda_{cv}$, which is defined as
\begin{equation}
\label{eq:supp_cost_volume_based_weight}
    \lambda_{cv} = max \left(\frac{|D_t^1 - D_t^{cv}|}{\delta_{D}}, 1 \right).
\end{equation}
Our regularization loss $L_{REG}$ is expressed as 
\begin{equation}
\label{eq:supp_reg_loss}
    L_{REG} = \lambda_{cv} \cdot max(|D_t^1 - D_t^2|, \mu_{cv} \delta_{D}),
\end{equation}
where $\mu_{cv} = [\lambda_{cv}=1]$ ($[\cdot]$ is an Iverson bracket) and $\delta_D$ is a hyperparamter of an allowable depth difference between $\theta_{depth}^2$ and $\theta_{depth}^{1*}$.
Note that small $\delta_{D}$ induces strong consistency while large $\delta_{D}$ alleviates the regularization strength.

In the last three rows of Table \ref{tab:supp_delta}, we present an analysis of various ${\delta_D}/{D_{max}}$ values within our regularization loss $L_{REG}$.
It is shown that our method maintains DE performance effectively across a range of $\delta_D$ settings, demonstrating the robustness of $L_{REG}$ against hyperparameter variations. 
Moreover, our regularization method offers a more suitable solution to regularizing the self-supervised monocular DE tasks.

\begin{table*}
\scriptsize
\begin{center}
    \begin{tabular}{cccccccccc}
\toprule
 & Methods & $abs\:rel\downarrow\cellcolor{c_lowbest}$  & $sq\:rel\downarrow\cellcolor{c_lowbest}$ & $rmse\downarrow\cellcolor{c_lowbest}$  & $log_{rmse}\downarrow\cellcolor{c_lowbest}$ & $a^1\uparrow\cellcolor{c_highbest}$ & $a^2\uparrow$\cellcolor{c_highbest}  & $a^3\uparrow$\cellcolor{c_highbest}  \\ 
\midrule
\multirow{14}{*}{\rotatebox[origin=c]{90}{Whole Image Region}} 
& Monodepth2 \cite{monodepth2} & 0.125 & 1.474 & 6.688 & 0.180 & 0.865 & 0.964 & 0.988 \\
& Ours-Monodepth2-C & 0.104   & 1.097  & 6.034  & 0.160  & 	0.895 & 0.972  & 0.99  \\
& Ours-Monodepth2 & 0.102  &   1.024  &   6.015  &   0.159  &   0.896  &   0.973  &   0.990  \\
\cmidrule{2-9}
& HR-Depth \cite{hr_depth} & 0.120  &   1.253  &   6.714  &   0.179  &   0.857  &   0.963  &   0.988  \\
& Ours-HR-Depth-C &  0.102  & 1.031  &   5.983  &   0.158  &   0.893  &   0.973  &   0.991 \\
& Ours-HR-Depth &  0.100  & 1.010  &   5.998  &   0.157  &   0.896  &   0.974  &   0.991 \\
\cmidrule{2-9}
& CADepth \cite{cadepth} & 0.124 & 1.278 & 6.771 & 0.183 & 0.862 & 0.962 & 0.986 \\
& Ours-CADepth-C &  0.099  &   1.018  &   5.706  &   0.152  &   0.903  &   0.977  &   0.992  \\
& Ours-CADepth &  0.097  &   0.966  &   5.646  &   0.150  &   0.907  &   0.978  &   0.992  \\
\cmidrule{2-9}
& MonoViT \cite{monovit} & 0.106 & 1.098 & 6.071 & 0.160 & 0.881 & 0.974 & 0.991 \\
& Ours-MonoViT-C & 0.089 & 0.826 & 5.494 & 0.142 & 0.911 & 0.980 & 0.993 \\
& Ours-MonoViT & 0.088  & 0.795 &  5.368  &  0.140  &  0.920 & 0.981  & 0.994 \\
\midrule
\multirow{14}{*}{\rotatebox[origin=c]{90}{Dynamic Object Region}}
& Monodepth2 \cite{monodepth2} & 0.185 & 2.432 & 5.919 & 0.218 & 0.794 & 0.918 & 0.962 \\
& Ours-Monodepth2-C & 0.125   & 1.185  & 4.459 & 0.161  & 	0.876 & 0.969  & 0.988  \\
& Ours-Monodepth2 & 0.119  &   1.044  &   4.270  &   0.157  &   0.881  &   0.970  &   0.989  \\
\cmidrule{2-9}
& HR-Depth \cite{hr_depth} & 0.165  &   2.144  &   5.720  &   0.198  &   0.811  &   0.940  &   0.974  \\
& Ours-HR-Depth-C &  0.110  & 1.177  &   4.601  &   0.160  &   0.878  &   0.972  &   0.989 \\
& Ours-HR-Depth &  0.113  & 1.015  &   4.297  &   0.154  &   0.883  &   0.974  &   0.991 \\
\cmidrule{2-9}
& CADepth \cite{cadepth} & 0.143 & 1.278 & 4.997 & 0.184 & 0.825 & 0.957 & 0.987 \\
& Ours-CADepth-C &  0.121  &   1.221  &   4.450  &   0.158  &   0.886  &   0.973  &   0.988  \\
& Ours-CADepth &  0.116  &   1.161  &   4.370  &   0.154  &   0.892  &   0.973  &   0.988  \\
\cmidrule{2-9}
& MonoViT \cite{monovit} & 0.149 & 1.508 & 5.340 & 0.190 & 0.817 & 0.945 & 0.983 \\
& Ours-MonoViT-C & 0.100 & 0.779 & 3.836 & 0.138 & 0.913 & 0.980 & 0.992 \\
& Ours-MonoViT & 0.098  & 0.674  & 3.688 &  0.135 &  0.914 & 0.981  &  0.992 \\
\bottomrule
\end{tabular}

\end{center}
\caption{Performance comparison of self-supervised DE methods trained with their original training strategy, our coarse training stage, and our coarse-to-fine training strategy on Cityscapes \cite{cityscapes} in a resolution of $192 \times 640$. For the DE performance in the region of objects in dynamic classes, we utilize masks provided by \cite{feng2022disentangling}.}
\label{tab:supp_cityscapes}
\end{table*}

\begin{table*}
\scriptsize
\begin{center}
    \begin{tabular}{cccccccccc}
\toprule
 & Methods & $abs\:rel\downarrow\cellcolor{c_lowbest}$  & $sq\:rel\downarrow\cellcolor{c_lowbest}$ & $rmse\downarrow\cellcolor{c_lowbest}$  & $log_{rmse}\downarrow\cellcolor{c_lowbest}$ & $a^1\uparrow\cellcolor{c_highbest}$ & $a^2\uparrow$\cellcolor{c_highbest}  & $a^3\uparrow$\cellcolor{c_highbest}  \\ 
\midrule
\multirow{14}{*}{\rotatebox[origin=c]{90}{Whole Image Region}} 
& Monodepth2 \cite{monodepth2} &  0.115 & 0.903 & 4.863 & 0.193 & 0.877 & 0.959 & 0.981 \\
& Ours-Monodepth2-C & 0.114   & 0.873  & 4.802  & 0.191  & 	0.877 & 0.960  & 0.981  \\
& Ours-Monodepth2 &  0.112  &   0.866  &   4.766  &   0.190  &   0.879  &   0.960  &   0.982  \\
\cmidrule{2-9}
& HR-Depth \cite{hr_depth} & 0.109 & 0.792 & 4.632 & 0.185 & 0.884 & 0.962 & 0.983 \\
& Ours-HR-Depth-C &  0.108  & 0.773  &   4.623  &   0.185  &   0.886  &   0.962  &   0.983 \\
& Ours-HR-Depth &  0.108 & 0.775 & 4.614 & 0.184 & 0.886 & 0.962 & 0.983 \\
\cmidrule{2-9}
& CADepth \cite{cadepth} & 0.105 & 0.769 & 4.535 & 0.181 & 0.892 & 0.964 & 0.983 \\
& Ours-CADepth-C &  0.104  &   0.742  &   4.481  &   0.179  &   0.895  &   0.966  &   0.984  \\
& Ours-CADepth &  0.103  &   0.730  &   4.427  &   0.179  &   0.895  &   0.966  &   0.984  \\
\cmidrule{2-9}
& MonoViT \cite{monovit} &  0.099 & 0.708 & 4.372 & 0.175 & 0.900 & 0.967 & 0.984 \\
& Ours-MonoViT-C & 0.097 & 0.743 & 4.418 & 0.173 & 0.903 & 0.968 & 0.984 \\
& Ours-MonoViT &  0.096  &   0.696  &   4.327  &  0.174  &  0.904  &   0.968  &   0.985  \\
\midrule
\multirow{14}{*}{\rotatebox[origin=c]{90}{Dynamic Object Region}}
& Monodepth2 \cite{monodepth2}  & 0.192  &   2.853  &   8.011  &   0.277  &   0.749  &   0.901  &   0.949 \\
& Ours-Monodepth2-C &    0.187  &   2.640  &   7.870  &   0.272  &   0.754  &   0.898  &   0.952    \\
& Ours-Monodepth2 & 0.183  &   2.539  &   7.723  &   0.268  &   0.757  &   0.900  &   0.954  \\
\cmidrule{2-9}
& HR-Depth \cite{hr_depth} & 0.191  &   2.722  &   7.859  &   0.273  &   0.743  &   0.890  &   0.949  \\
& Ours-HR-Depth-C &  0.179  & 2.249  &   7.439  &   0.264  &   0.753  &   0.907  &   0.955 \\
& Ours-HR-Depth &  0.177  & 2.191  &   7.395  &   0.264  &   0.755  &   0.907  &   0.955 \\
\cmidrule{2-9}
& CADepth \cite{cadepth} & 0.174 & 2.208 & 7.361 & 0.263 & 0.764 & 0.902 & 0.954 \\
& Ours-CADepth-C &  0.166  &   1.884  &   7.002  &   0.254  &   0.765  &   0.914  &   0.959  \\
& Ours-CADepth & 0.164  &   1.794  &   6.873  &   0.250  &   0.767  &   0.920  &   0.963  \\
\cmidrule{2-9}
& MonoViT \cite{monovit}  & 0.161	& 1.944	& 7.032	& 0.250	& 0.790	& 0.921	& 0.959 \\
& Ours-MonoViT-C & 0.157 & 1.716 & 6.796 & 0.244 & 0.797 & 0.927 & 0.963 \\
& Ours-MonoViT & 0.155  & 1.659  & 6.752 &  0.244 &  0.796 & 0.924  &  0.963 \\
\bottomrule
\end{tabular}

\end{center}
\caption{Performance comparison of self-supervised DE methods trained with their original training strategy, our coarse training stage, and our coarse-to-fine training strategy on KITTI \cite{kitti2015} in a resolution of $192 \times 640$. For the DE performance in the region of objects in dynamic classes, we utilize Mask2Former \cite{cheng2021mask2former} instance segmentation network to obtain binary instance masks that include cars, bicycles, and pedestrians.}
\label{tab:supp_kitti}
\end{table*}

\vspace{2mm}
\section{Additional Experimental Results}

In Table \ref{tab:supp_cityscapes} and \ref{tab:supp_kitti}, we provide the DE performance comparison between existing self-supervised monocular DE methods including Monodepth2 \cite{monodepth2}, HR-Depth \cite{hr_depth}, CADepth \cite{cadepth}, MonoViT \cite{monovit} and those trained with our coarse-to-fine training strategy on Cityscapes \cite{cityscapes} and KITTI \cite{kitti2015}, respectively.
We denote each model trained with our coarse-training stage as Ours-Monodepth2-C, Ours-HR-Depth-C, Ours-CADepth-C, and Ours-MonoViT-C.
Also, we denote each model trained with our full coarse-to-fine training strategy as Ours-Monodepth2, Ours-HR-Depth, Ours-CADepth, and Ours-MonoViT, respectively.

\vspace{1mm}
\subsection{Cityscapes Results}
\label{sec:supp_cityscapes_results}

In Table \ref{tab:supp_cityscapes}, it is shown that our coarse-to-fine training strategy consistently enhances the DE performance of various models \cite{monodepth2, hr_depth, cadepth, monovit}.
This enhancement is evident in both overall image regions and specifically within the region of objects in dynamic classes. 
A notable performance enhancement is observed when comparing models trained with traditional methods against those utilizing our coarse stage, where the Ground-contacting-prior Disparity Smoothness Loss (GDS-Loss) ensures precise depth supervision, especially for dynamic objects.
Also, it should be noted that our fine training stage further enhances the DE performance of the models.
This indicates the effectiveness of our coarse-to-fine training strategy in handling the moving object problem in self-supervised monocular depth estimation, seamlessly integrating the existing DE methods without the need for modifications such as auxiliary object motion estimation networks.

\vspace{1mm}
\subsection{KITTI Results}
\label{sec:supp_kitti_results}

Table \ref{tab:supp_kitti} presents a comprehensive evaluation of our coarse-to-fine training strategy applied to the existing DE methods \cite{monodepth2, hr_depth, cadepth, monovit} on KITTI \cite{kitti2015}.
Although KITTI dataset \cite{kitti2015} contains a relatively smaller amount of moving objects compared to Cityscapes \cite{cityscapes}, our training strategy consistently enhances DE performance across all models, particularly in regions of objects in dynamic classes.
This consistent improvement shows the generalizability and effectiveness of our self-supervised monocular depth estimation training strategy in outdoor monocular driving datasets. 

\begin{figure*}
    \centerline{\includegraphics[width=1\textwidth]{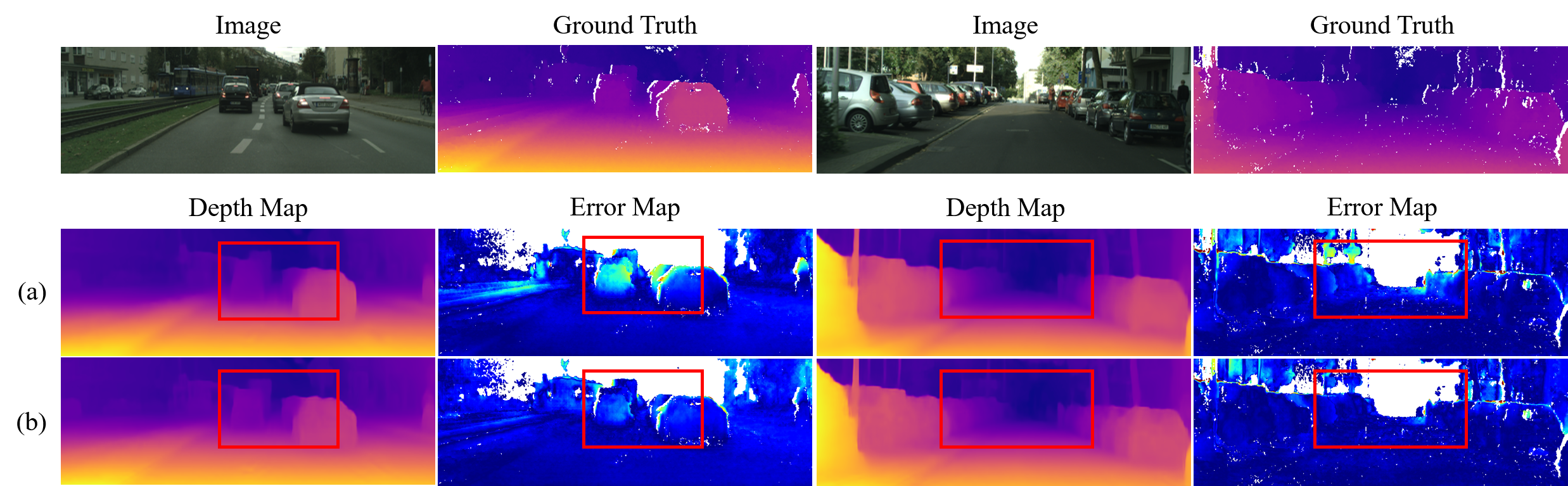}}
    \caption{Performance comparison on estimated depth maps and error maps between (a) Ours-MonoViT-C, trained with the coarse stage, and (b) Ours-MonoViT, trained with our full training strategy, in Cityscapes \cite{cityscapes}.} 
    \label{fig:supp_coarse_to_fine}
\end{figure*}

\vspace{3mm}
\subsection{Qualitative Results}
\label{sec:supp_qualitative_results}

We provide additional qualitative results to highlight the effectiveness of our coarse-to-fine training strategy in enhancing depth estimation (DE) performance.

\subsubsection{Effectiveness of Our Fine Training Stage}

In Figure \ref{fig:supp_coarse_to_fine}, we demonstrate the impact of our fine training strategy with the proposed regularization loss.
Although our Ground-contacting-prior Disparity Smoothness Loss (GDS-Loss) induces a DE network to predict the consistent depth of objects and their bottom region, it can occasionally inaccurately estimate the depth of objects when their depth varies in the vertical direction or their bottom parts are occluded by other objects.
As shown in the second row of Figure \ref{fig:supp_coarse_to_fine}, the MonoViT \cite{monovit} trained with our coarse training stage predicts the comparably imprecise depth of the cars in the red boxes.
On the other hand, Ours-MonoViT, benefiting from our fine training stage, achieves more precise depth on the cars in the red boxes.
Under the carefully designed regularization loss, the DE network learned to estimate detailed depth of the objects in dynamic classes.
The enhancements from our fine training stage are clearly shown in the error maps. 
This improvement highlights the capability of our training strategy to achieve not only accurate but also detailed depth estimations for objects.

\subsubsection{Depth Map Comparisons with Existing Methods}
\begin{figure*}
    \centerline{\includegraphics[width=1\textwidth]{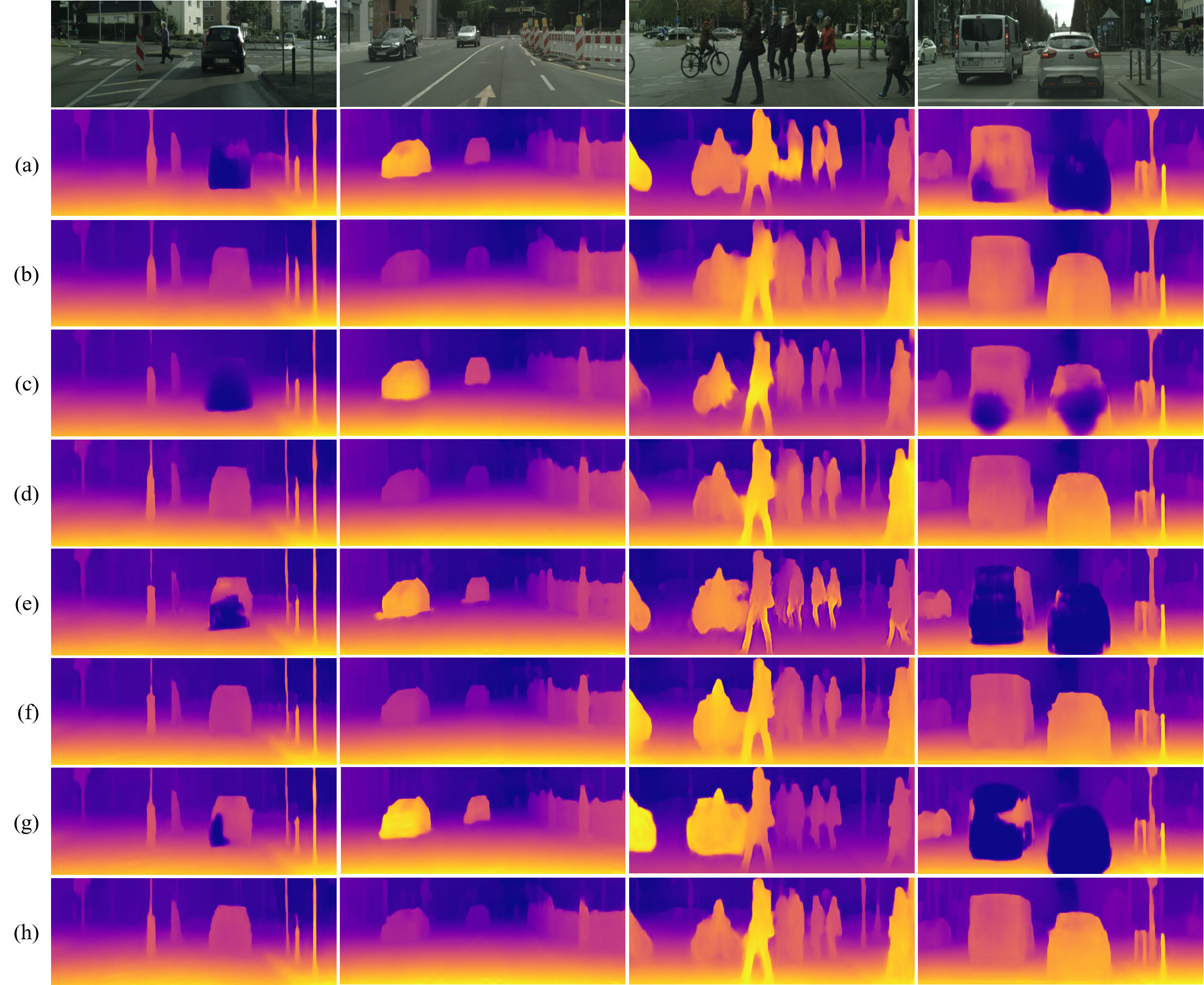}}
    \caption{Performance comparison on estimated depth maps between (a) Monodepth2 \cite{monodepth2}, (b) Ours-Monodepth2, (c) HR-Depth \cite{hr_depth}, (d) Ours-HR-Depth, (e) CADepth \cite{cadepth}, (f) Ours-CADepth, (g) MonoViT \cite{monovit} and (h) Ours-MonoViT in Cityscapes \cite{cityscapes}.} 
    \label{fig:supp_cityscapes}
\end{figure*}

Figure \ref{fig:supp_cityscapes} shows the predicted depth maps from the existing self-supervised DE methods \cite{monodepth2, hr_depth, cadepth, monovit} and those models trained with our training strategy.
As shown, the original models predict erroneous depth, especially in the region of objects in dynamic classes such as cars, bicycles, and pedestrians.
The original models predict inaccurate depth on the object regions, since the objects are not excluded from the calculation of the reprojection loss and the supervision of their depth is insufficient in the conventional self-supervised monocular depth learning pipeline.
In contrast, our training strategy with precise depth supervision on the object regions by using our GDS-Loss yields precise depth estimation.
It is noteworthy that our training strategy is easily integrated into various network architectures and boosts their DE performance, handling moving object problems. 

\subsubsection{3D Point Cloud Reconstructions: MonoViT vs. Ours-MonoViT}
\begin{figure*}
    \centerline{\includegraphics[width=1\textwidth]{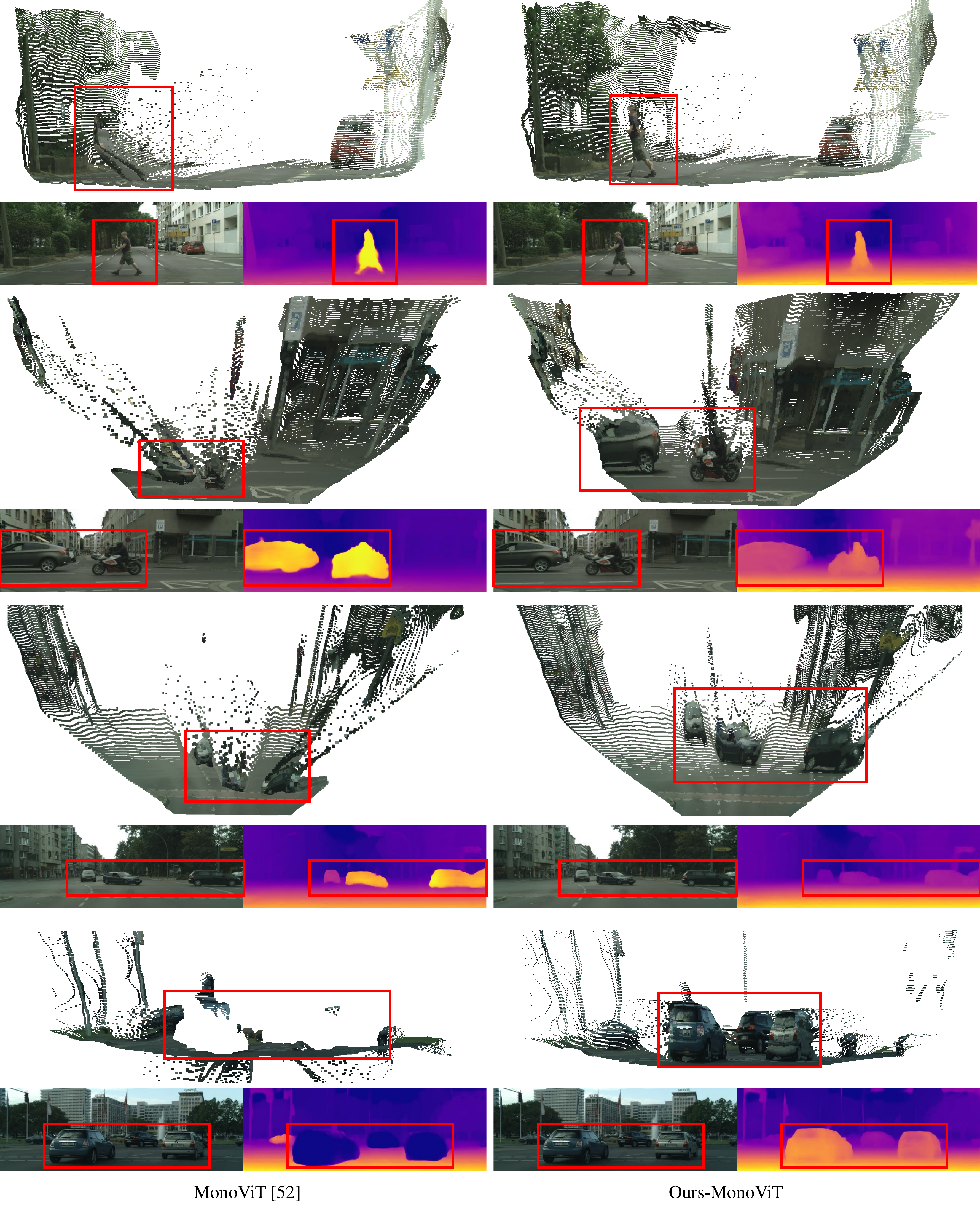}}
    \caption{Performance comparison on estimated depth maps and the snapshots of 3D reconstructed point clouds between MonoViT \cite{monovit} and Ours-MonoViT in Cityscapes \cite{cityscapes}.} 
    \label{fig:supp_point_cloud}
\end{figure*}

Figure \ref{fig:supp_point_cloud} displays the estimated depth maps and the reconstructed point clouds from MonoViT \cite{monovit} and Ours-MonoViT.
As MonoViT \cite{monovit} struggles to estimate accurate depth on the object regions, the objects in the reconstructed point clouds are floating or sunken under the ground. 
On the other hand, Ours-MonoViT predicts the accurate depth of objects in dynamic classes such as cars, motorcycles, and pedestrians.
Thus, the objects are standing on the ground in the reconstructed point clouds.

\end{document}